\documentclass[twocolumn,10pt]{asme2ej}
\newcommand{\EQ}{\begin{equation}}
\newcommand{\NQ}{\end{equation}}
\newcommand{\ER}{\begin{eqnarray}}
\newcommand{\NR}{\end{eqnarray}}
\newcommand{\ERS}{\begin{eqnarray*}}
\newcommand{\NRS}{\end{eqnarray*}}
\newcommand{\bit}{\begin{itemize}}
\newcommand{\ben}{\begin{enumerate}}
\newcommand{\eben}{\end{enumerate}}
\newcommand{\ebit}{\end{itemize}}
\newcommand{\bzero}{{\bf 0}}

\newcommand{\bh}{{\bf h}}

\newcommand{\bm}{{\bf m}}

\newcommand{\bq}{{\bf q}}

\newcommand{\bu}{{\bf u}}

\newcommand{\bx}{{\bf x}}
\newcommand{\by}{{\bf y}}

\newcommand{\bA}{{\bf A}}
\newcommand{\bB}{{\bf B}}

\newcommand{\bK}{{\bf K}}

\newcommand{\bM}{{\bf M}}

\newcommand{\bP}{{\bf P}}
\newcommand{\bQ}{{\bf Q}}

\newcommand{\dq}{\dot q}

\def\m@th{\mathsurround=0pt }

\def\eqalign#1{\null\,\vcenter{\openup1\jot \m@th
  \ialign{\strut\hfil$\displaystyle{##}$&$\displaystyle{{}##}$\hfil
      \crcr#1\crcr}}\,}

\def\displ@y{\global\dt@ptrue \openup1\jot \m@th
  \everycr{\noalign{\ifdt@p \global\dt@pfalse
      \vskip-\lineskiplimit \vskip\normallineskiplimit
      \else \penalty\interdisplaylinepenalty \fi}}}
\def\@lign{\tabskip=0pt\everycr={}} 

\def\eqalignno#1{\displ@y \tabskip=\centering
  \halign to\displaywidth{\hfil$\@lign\displaystyle{##}$\tabskip=0pt
    &$\@lign\displaystyle{{}##}$\hfil\tabskip=\centering
    &\llap{$\@lign##$}\tabskip=0pt\crcr
    #1\crcr}}

\def\oalpha{{\ooalign{\hfil\raise.07ex\hbox{$\alpha$}\hfil\crcr\mathhexbox20D}}}

\def\ooalpha{{\ooalign{\hfil\raise.07ex\hbox{$\alpha$}\hfil\crcr$\textstyle{\circ}t$}}}

\def\obeta{{\ooalign{\hfil\raise.07ex\hbox{$\scriptstyle{\beta}$}\hfil\crcr$\textstyle{\circ}$}}}

\def\ot{{\ooalign{\hfil\raise.07ex\hbox{$t$}\hfil\crcr$\textstyle{\circ}$}}}

\def\oooalpha{{\ooalign{\hfil\raise.07ex\hbox{${\scriptstyle{\alpha}}$}\hfil\crcr${\textstyle{\circ}}$}}}

\def\trademark{{\ooalign{\hfil\raise.07ex\hbox{R}\hfil\crcr\mathhexbox20D}}}

\usepackage{graphicx} 
\usepackage{hyperref}   
\hypersetup{
	colorlinks=true,
	linkcolor=blue,
	citecolor=blue,
	urlcolor=blue,
}
\usepackage[square,numbers]{natbib}

\usepackage{amsbsy,latexsym,mathrsfs,mathtools}
\usepackage{mathptmx} 
\DeclareMathAlphabet{\mathcal}{OMS}{cmsy}{m}{n}
\usepackage{bm}
\usepackage{float}

\usepackage{times,color,soul} 
\usepackage{amssymb}
\usepackage{color,soul}
\usepackage{dsfont}
\usepackage{comment}
\usepackage{algorithm}
\usepackage{algorithmic}
\usepackage{psfrag}   
\usepackage{multirow}
\usepackage{url}
\usepackage{hyperref}
\usepackage{setspace}

\usepackage[table]{xcolor}
\usepackage{fancyhdr}
\usepackage{paralist}
\usepackage{booktabs}
\usepackage{gensymb}
\usepackage{lineno}
\usepackage{flushend}

\newtheorem{thm}{Theorem}

\newtheorem{prop}{Proposition}

\title{Provably Stabilizing Global-Position Tracking Control for Hybrid Models of Multi-Domain Bipedal Walking via Multiple Lyapunov Analysis}

\author{Yuan Gao \thanks{Equal contribution.}
    \affiliation{
	Dept. of Mechanical Engineering\\
	University of Massachusetts\\
	Lowell, Massachusetts, 01854\\
    Email: Yuan\_Gao@student.uml.edu
    }
}

\author{Kentaro Barhydt$~{}^*$
    \affiliation{ 
	Department of Mechanical Engineering\\
    Massachusetts Institute of Technology
    \\
    Cambridge, Massachusetts, 02139\\
        Email: kbarhydt@mit.edu
    }
}

\author{Christopher Niezrecki
    \affiliation{
        Department of Mechanical Engineering\\
        The University of Massachusetts\\
        Lowell, Massachusetts, 01851\\
Christopher\_Niezrecki@uml.edu
    }
}

\author{Yan Gu\thanks{Address all correspondence to this author. This material is based upon work supported by the National Science Foundation under Grant No. CMMI-1934280.}
    \affiliation{
    School of Mechanical Engineering\\
	Purdue University\\
	West Lafayette, Indiana, 47907\\
        Email: yangu@purdue.edu
    }
}

\begin{document}

\maketitle

\begin{abstract}
\textit{Accurate control of a humanoid robot's global position (i.e., its three-dimensional position in the world) is critical to the reliable execution of high-risk tasks such as avoiding collision with pedestrians in a crowded environment.
This paper introduces a time-based nonlinear control method that achieves accurate global-position tracking (GPT) for multi-domain bipedal walking.
Deriving a tracking controller for bipedal robots is challenging due to the highly complex robot dynamics that are time-varying and hybrid, especially for multi-domain walking that involves multiple phases/domains of full actuation, over actuation, and underactuation.
To tackle this challenge, we introduce a continuous-phase GPT control law for multi-domain walking, which provably ensures the exponential convergence of the entire error state within the full and over actuation domains and that of the directly regulated error state within the underactuation domain.
We then construct sufficient multiple-Lyapunov stability conditions for the hybrid multi-domain tracking error system under the proposed GPT control law.
We illustrate the proposed controller design through both three-domain walking with all motors activated and two-domain gait with inactive ankle motors.
Simulations of a ROBOTIS OP3 bipedal humanoid robot demonstrate the satisfactory accuracy and convergence rate of the proposed control approach under two different cases of multi-domain walking as well as various walking speeds and desired paths.}
\end{abstract}


\section{Introduction}

Multi-domain walking of legged locomotors refers to the type of walking that involves multiple continuous foot-swinging phases and discrete foot-landing behaviors within a gait cycle, due to changes in foot-ground contact conditions and actuation authority
~\cite{zhao2017multi,zhao2014human}. 
Human walking is a multi-domain process that involves phases with different actuation types.
These phases include: (1) full actuation phases during which the support foot is flat on the ground and the number of actuators is equal to that of the degrees of freedom (DOFs); (2) underactuation phases where the support foot rolls about its toe and the number of actuators is less than that of the DOFs; and (3) over actuation phases within which both feet are on the ground and there are more actuators than DOFs.

\begin{figure}[t]
	\centering
	\includegraphics[width=0.9\linewidth]{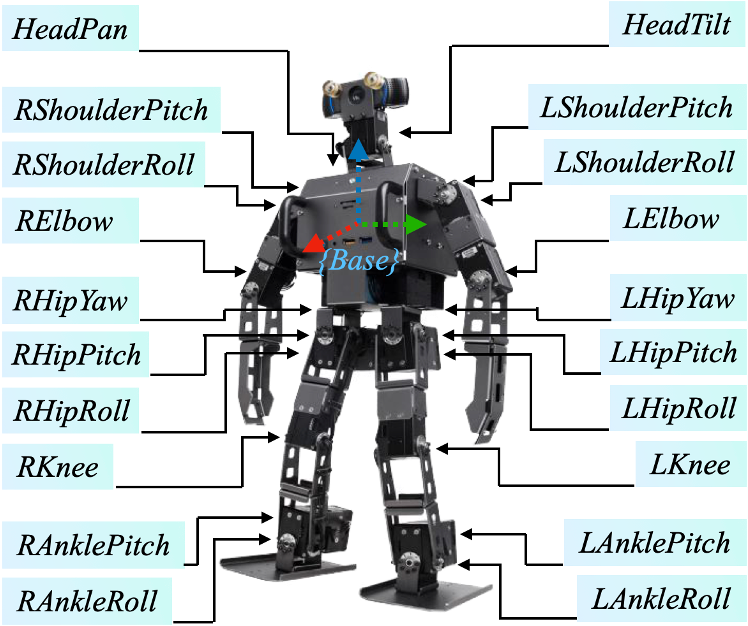}
\caption{
Illustration of the Darwin OP3 robot, which is used to validate the proposed global-position tracking control approach.
 Darwin OP3 is a bipedal humanoid robot with twenty revolute joints, designed and manufactured by ROBOTIS~\cite{robotis}.
The reference frame of the robot's floating base, highlighted as ``$\{ \text{Base}\}$'', is located at the center of the chest.}
	\label{fig: op3 joints illustration}
\end{figure}

Researchers have proposed various control strategies to achieve stable multi-domain walking for bipedal robots.
Zhao et al.\cite{zhao2014human} proposed a hybrid model to capture the multi-domain robot dynamics and used offline optimization to obtain the desired motion trajectory based on the hybrid model.
An input-output linearizing control scheme was then applied to drive the robot state to converge to the desired trajectory.
The approach was validated on a physical planar bipedal robot, AMBER2, and later extended to another biped platform~\cite{zhao2017multi}, ATRIAS~\cite{Ramezani2014}.
Hereid et al. utilized the reduced-order Spring Loaded Inverted Pendulum model \cite{schwind1998spring} to design an optimization-based trajectory generation method that plans periodic orbits in the state space of the compliant bipedal robot~\cite{hereid2014dynamic}, ATRIAS~\cite{grimes2012design}.
The method guarantees orbital stability of the multi-domain gait based on the hybrid zero dynamics (HZD) approach \cite{westervelt2007feedback}.
Reher et al. achieved an energy-optimal multi-domain walking gait on the physical robot platform, DURUS, by creating a hierarchical motion planning and control framework~\cite{reher2016realizing}.
The framework ensures orbital walking stability and energy efficiency for the multi-domain robot model based on the HZD approach~\cite{westervelt2007feedback}.
Hamed et al. established orbitally stable multi-domain walking on a quadrupedal robot~\cite{hamed2019dynamically,akbari2019dynamic} by modeling the associated hybrid full-order robot dynamics and constructing virtual constraints~\cite{grizzle2001asymptotically}.
Although these approaches have realized provable stability and impressive performance of multi-domain walking on various physical robot platforms, it remains unclear how to directly extend them to solve general global-position tracking (GPT) control problems.
In real-world mobility tasks, such as dynamic obstacle avoidance during navigation through a crowded hallway, a robot needs to control its global position accurately with precise timing.
However, the previous methods' walking stabilization mechanism is orbital stabilization \cite{westervelt2003hybrid,westervelt2007feedback,sreenath2011compliant,veer2019input,gong2019feedback}, which may not ensure reliable tracking of a time trajectory precisely with the desired timing.

We have developed a GPT control method that achieves exponential trajectory tracking for the hybrid model of two-dimensional (2-D) fully actuated bipedal walking~\cite{gu2016bipedal,gu2018straight,gu2017time}.
To extend our approach to 3-D fully actuated robots, we considered the robot's lateral global movement and its coupling with forward dynamics through dynamics modeling and stability analysis~\cite{gao2019global,gu2020adaptive,gu2021adaptive,gu2022global}.
For fully actuated quadrupedal robotic walking on a rigid surface moving in the inertial frame, we formulated the associated robot dynamics as a hybrid time-varying system and exploited the model to develop a GPT control law for fully actuated quadrupeds~\cite{iqbaloptimization,9108552,iqbal2023real}.
However, these methods designed for fully actuated robots cannot solve the multi-domain control problem directly because they do not explicitly handle the underactuated robot dynamics associated with general multi-domain walking.

Some of the results presented in this paper have been reported in~\cite{gao_multi_2019global}.
While our previous work in~\cite{gao_multi_2019global} focused on GPT controller design and stability analysis for hybrid multi-domain models of 2-D walking along a straight line, this study extends the previous method to 3-D bipedal robotic walking, introducing the following significant new contributions:
\begin{itemize}
\item [(a)] Theoretical extension of the previous GPT control method from 2-D to 3-D bipedal robotic walking. The key novelty is the formulation of a new phase variable that represents the distance traveled along a general curved walking path and can be used to encode the desired global-position trajectories along both straight lines and curved paths.
\item [(b)] Lyapunov-based stability analysis to generate sufficient conditions under which the proposed GPT control method provably stabilizes 3-D multi-domain walking. Full proofs associated with the stability analysis are provided, while only sketches of partial proofs were reported in~\cite{gao_multi_2019global}.
\item [(c)] 
Extension from three-domain walking with all motors activated to two-domain gait with inactive ankle motors, by formulating a hybrid two-domain system and developing a GPT controller for this new gait type. Such an extension was missing in~\cite{gao_multi_2019global}.
\item [(d)] Validation of the proposed control approach through MATLAB simulations of a ROBOTIS OP3 humanoid robot (see Fig.\ref{fig: op3 joints illustration}) with different types of multi-domain walking, both straight and curved paths, and various desired global-position profiles. In contrast, our previous validation only used a simple 2-D biped with seven links~\cite{gao_multi_2019global}.
\item [(e)] Casting the multi-domain control law as a quadratic program (QP) to ensure the feasibility of joint torque limits, and comparing its performance with an input-output linearizing control law, which were not included in~\cite{gao_multi_2019global}.
\end{itemize}

This paper is structured as follows.
Section~\ref{Section-Dyanmics} explains the full-order robot dynamics model associated with a common three-domain walking gait.
Section~\ref{Section-Control} presents the proposed GPT control law for three-domain walking.
Section~\ref{Section-Stability} introduces the Lyapunov-based closed-loop stability analysis. 
Section~\ref{Section-Second MD walking} summarizes the controller design extension from three-domain walking to a two-domain gait.
Section~\ref{Section-Simulation} reports the simulation validation results.
Section~\ref{Sec: Discussion} discusses the capabilities and limitations of the proposed control approach.
Section~\ref{Sec: Conclusion} provides the concluding remarks.
Proofs of all theorems and propositions are given in Appendix~\ref{proof}.

 \begin{figure}[t]
	\centering
	\includegraphics[width=0.7\linewidth]{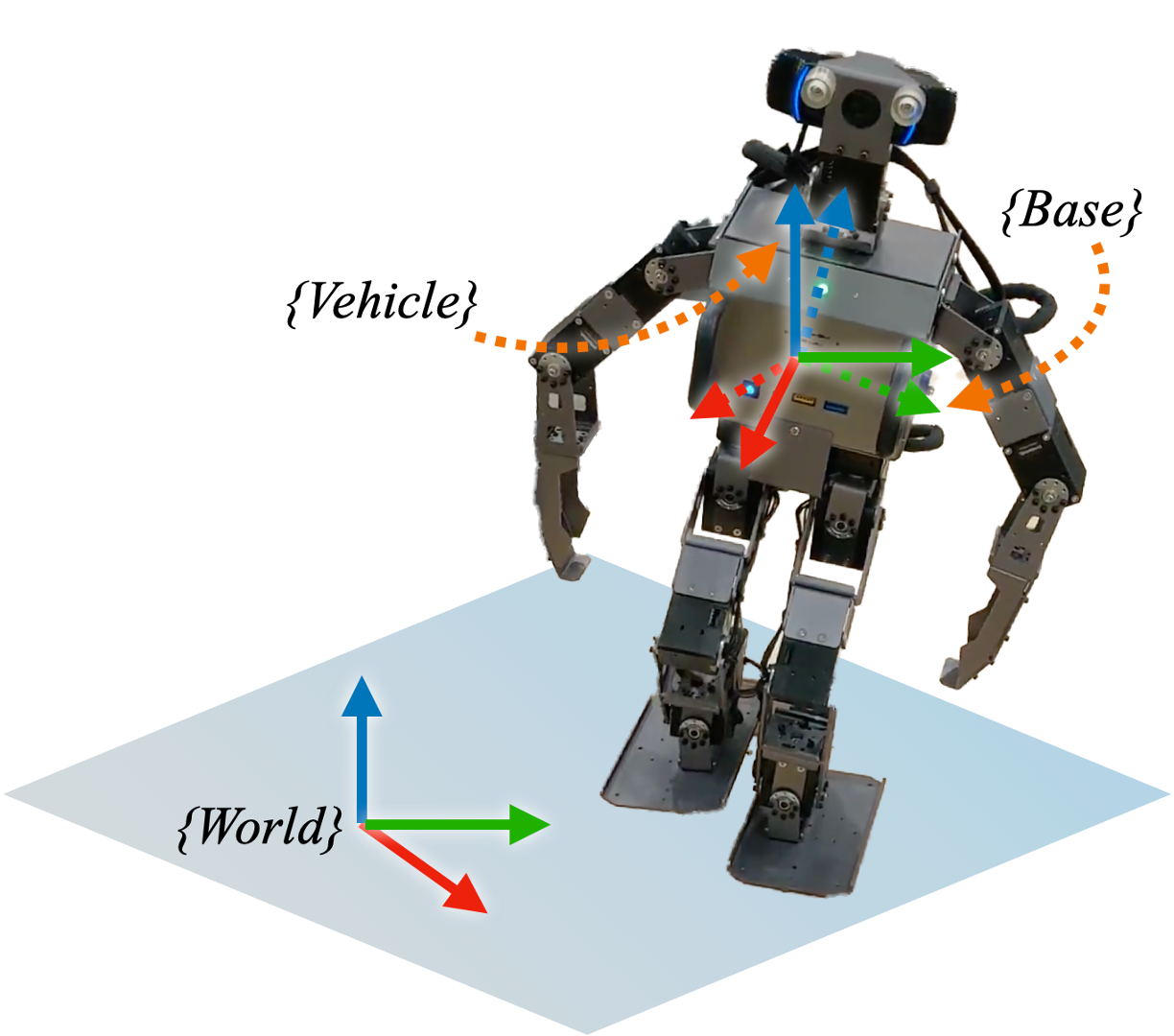}
	\caption{
	Illustration of the three coordinate systems used in the study: world frame, vehicle frame, and base frame.}
	\label{fig: coordinate system}
\end{figure}

\section{FULL-ORDER DYNAMIC MODELING OF THREE-DOMAIN WALKING}
\label{Section-Dyanmics}
This section presents the hybrid model of bipedal robot dynamics associated with three-domain walking.

\subsection{Coordinate Systems and Generalized Coordinates}

This subsection explains the three coordinate systems used in the proposed controller design.
Figure~\ref{fig: coordinate system} illustrates the three frames, with the $x$-, $y$-, and $z$-axes respectively highlighted in red, green, and blue.

\vspace{-0.2in}
\subsubsection{World frame}
The world frame, also known as the inertial frame, is rigidly attached to the ground (see ``\{World\}'' in Fig.~\ref{fig: coordinate system}).

\vspace{-0.2in}
\subsubsection{Base frame}
The base frame, illustrated as ``\{Base\}'' in Fig.~\ref{fig: coordinate system}, is rigidly attached to the robot's trunk. The $x$-direction (red) points forward, and the $z$-direction (blue) points towards the robot's head. 

\vspace{-0.2in}
\subsubsection{Vehicle frame}
The origin of the vehicle frame (see ``\{Vehicle\}'' in Fig.~\ref{fig: coordinate system}) coincides with the base frame, and its $z$-axis remains parallel to that of the world frame.
The vehicle frame rotates only about its z-axis by a certain heading (yaw) angle. 
The yaw angle of the vehicle frame with respect to (w.r.t) the world frame equals that of the base frame w.r.t. the world frame, while the roll and pitch angles of the vehicle frame w.r.t the world frame are 0.

\vspace{-0.2in}
\subsubsection{Generalized coordinates}
To use Lagrange's method to derive the robot dynamics model, we need to first introduce the generalized coordinates to represent the base pose and joint angles of the robot.

We use $\mathbf{p}_b \in \mathbb{R}^3$ and $\boldsymbol{\gamma}_b\in SO(3)$
to respectively denote the absolute base position and orientation w.r.t the world frame, and their coordinates are represented by
($x_b,y_b,z_b$) and 
 ($\phi_b,\theta_b,\psi_b$). 
 Here $\phi_b,\theta_b,\psi_b$ are the roll, pitch, and yaw angles, respectively.
Then, the 6-D pose $\mathbf{q}_b$ of the base is given by: $\mathbf{q}_b:=[\mathbf{p}_b^T,\boldsymbol{\gamma}_b^T]^T$.

Let the scalar real variables
$q_1,~...,~q_{n}$
represent the joint angles of the $n$ revolute joints of the robot. 
Then, the generalized coordinates of a 3-D robot, which has a floating base and $n$ independent revolute joints, can be expressed as:
\vspace{-0.2in}
\begin{equation}
\vspace{-0.2in}
\label{equ:generalized coordinate}
\mathbf{q}=
\begin{bmatrix}
    \mathbf{q}_b^T,~q_1,~...,~q_{n}
\end{bmatrix}
    ^T
    \in \mathcal{Q},
\end{equation}
where $\mathcal{Q} \subset \mathbb{R}^{n+6}$ is the configuration space.
Note that the number of degrees of freedom (DOFs) of this robot without subjecting to any holonomic constraints is $n+6$.

\subsection{Walking Domain Description}

\label{Sec: walking domain}

For simplicity and without loss of generality, we consider the following assumptions on the foot-ground contact conditions during 3-D walking:
\begin{itemize}
    \item[(A1)]
    The toe and heel are the only parts of a support foot that can contact the ground~\cite{zhao2017multi}.
    \item[(A2)] While contacting the ground, the toes and/or heels have line contact with the ground.
    \item [(A3)] There is no foot slipping on the ground. 
\end{itemize}

Also, we consider the common assumption below about the robot's actuators:
\begin{itemize}
    \item [(A4)] All the $n$ revolute joints of the robot are independently actuated.
\end{itemize}
Let $n_a$ denote the number of independent actuators, and $n_a=n$ holds under assumption (A4).

\begin{figure}[t]
	\centering
	\includegraphics[width=1\linewidth]{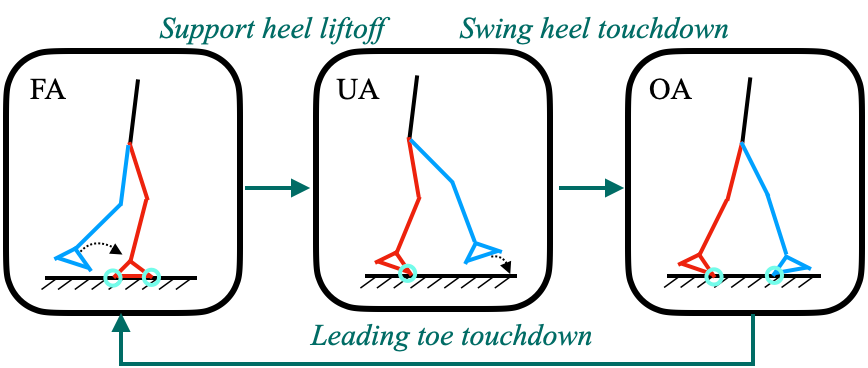}
    \vspace{-0.2in}
	\caption{The directed cycle of 3-D three-domain walking. 
 The green circles in the diagram highlight the portions of a foot that are in contact with the ground.
 The position trajectory of the swing foot is indicated by the dashed arrow.
 The red and blue legs respectively represent the support and swing legs. Note that when the robot exits the OA domain and enters the FA domain, the swing and support legs switch their roles, and accordingly the leading and trailing legs swap their colors.}
	\label{fig: multi_domain_walking1}
    \end{figure}

Figure \ref{fig: multi_domain_walking1} illustrates the complete gait cycle of human-like walking with a rolling support foot.
As the figure displays, the complete walking cycle involves three continuous phases/domains and three discrete behaviors connecting the three domains.
The three domains are: 
\begin{itemize}
    \item [(i)] Full actuation (FA) domain, where $n_a$ equals the number of DOFs; 
    \item [(ii)] Underactation (UA) domain, where the number of independent actuators ($n_a$) is less than that of the robot's DOFs;
    and 
    \item [(iii)] Over actuation (OA) domain, where $n_a$ is greater than the number of DOFs.
\end{itemize}

The actuation types associated with the three domains are different because those domains have distinct foot-ground contact conditions, which are explained next under assumptions (A1)-(A4).

    \subsubsection{FA domain}
    As illustrated in the ``FA'' portion of Fig.~\ref{fig: multi_domain_walking1}, only one foot is in support and it is static on the ground within the FA domain.
    Under assumption (A1), we know both the toe and heel of the support foot contact the ground.
    From assumptions (A2) and (A3), we can completely characterize the foot-ground contact condition with six independent scalar holonomic constraints.
    Using $n_c$ to denote the number of holonomic constraints, we have $n_c=6$ within an FA domain, and the number of DOFs becomes $\text{DOF}=n+6-n_c=n$.
    Meanwhile, $n_a=n$ holds under assumption (A4).
    Since $\text{DOF}=n_a$, all of the DOFs are directly actuated; that is, the robot is indeed fully actuated.

    \subsubsection{UA domain}
    The ``UA'' portion of Fig.~\ref{fig: multi_domain_walking1} shows that the robot's support foot rolls about its toe within a UA domain. 
    Under assumptions (A2) and (A3), the number of holonomic constraints is five, i.e., $n_c=5$.
    This is because the support foot can only roll about the line toe but its motion is fully restricted in terms of the 3-D translation and the pitch and yaw rotation. 
    Then, the number of DOFs is: $\text{DOF} = n + 6 - 5 = n+1$.
    Since the number of independent actuators, $n_a$, equals $n$ under assumption (A4) and is lower than the number of DOFs, $(n+1)$, the robot is underactuated with one degree of underactuation.

\subsubsection{OA domain}
   Upon exiting the UA domain, the robot's swing-foot heel strikes the ground and enters the OA domain (Fig.~\ref{fig: multi_domain_walking1}).
   Within an OA domain, both the trailing toe and the leading heel of the robot contact the ground, which is described by ten scalar holonomic constraints (i.e., $n_c=10$).
   Thus, the DOF becomes $ \text{DOF} = n + 6 - n_c = n-4$, which is less than the number of actuators under assumption (A4), meaning the robot is over actuated.

\subsection{Hybrid Multi-Domain Dynamics}

This subsection presents the full-order model of the robot dynamics that corresponds to multi-domain walking. Since multi-domain walking involves both continuous-time dynamics and discrete-time behaviors, a hybrid model is employed to describe the robot dynamics. To aid the readers in comprehending the hybrid system, the fundamentals of hybrid systems will be discussed first.

\subsubsection{Preliminaries on hybrid systems}
A hybrid control system $\mathcal{HC}$ is a tuple:
\vspace{-0.2in}
\begin{equation*}
\vspace{-0.2in}
    \mathcal{HC} = (\Gamma, D, U, S, \Delta, FG),
\end{equation*}
where
\begin{itemize}
    \item[${\blacksquare}$]
    The oriented graph $\Gamma=(V,E)$ comprises a set of vertices $V=\{v_1,v_2,...,v_N \}$ and a set of edges $E=\{e_1,e_2,...,e_N \}$, where $N$ is the total number of elements in each set.
    In this paper, each vertex $v_i$ represents the $i^{th}$ domain, while each edge $e_i$ represents the transition from the source domain to the target domain, thereby indicating the ordered sequence of all domains.
    For three-domain walking, we have $i=3$.
    \item[$\blacksquare$] $D$ is a set of domains of admissibility, which are the FA, UA, and OA domains for three-domain walking.
    \item[$\blacksquare$] $U$ is the set of admissible control inputs.
    \item[$\blacksquare$] $S$ is a set of switching surfaces determining the occurrence of switching between domains.
    \item[$\blacksquare$] $\Delta$ is a set of reset maps, which represents the impact dynamics between a robot's swing foot and the ground.
    \item[$\blacksquare$] $FG$ is a set of vector fields on the state manifold.
\end{itemize}
The elements of these sets are explained next.

\subsubsection{Continuous-phase dynamics}
Within any of the three domains, the robot only exhibits continuous movements, and its dynamics model is naturally continuous-time.~Applying Lagrange's method, we obtain the second-order, nonlinear robot dynamics as:
\vspace{-0.2in}
\begin{equation} 
\vspace{-0.1in}
\label{equ: dynamics}
\mathbf{M(q)}\ddot{\mathbf{q}}+\mathbf{c}(\mathbf{q},\dot{\mathbf{q}}) = \mathbf{B}\mathbf{u}+\mathbf{J}^T\mathbf{F}_{c},
\end{equation}
where $\mathbf{M}(\mathbf{q}):\mathcal{Q}\rightarrow\mathbb{R}^{(n+6)\times(n+6)}$ is the inertia matrix.
The vector $\mathbf{c}: \mathcal{TQ}\rightarrow\mathbb{R}^{(n+6)}$ is the summation of the Coriolis, centrifugal, and gravitational terms, where $\mathcal{TQ}$ is the tangent bundle of $\mathcal{Q}$.
The matrix $\mathbf{B}\in\mathbb{R}^{(n+6)\times n_a}$ is the input matrix.
The vector $\mathbf{u} \in {U} \subset\mathbb{R}^{n_a}$ is the joint torque vector.
The matrix $\mathbf{J}(\bq):\mathcal{Q}\rightarrow \mathbb{R}^{n_{c}\times (n+6)}$ represents the Jacobian matrix.
The vector $\mathbf{F}_c\in\mathbb{R}^{n_c}$ is the constraint force that the ground applies to the foot-ground contact region of the robot.
Note that the dimensions of $\mathbf{J}$ and $\mathbf{F}_c$ vary among the three domains due to differences in the ground-contact conditions.

The holonomic constraints can be expressed as:
\vspace{-0.2in}
\begin{equation} 
\vspace{-0.2in}
\label{equ: holonomic constraint}	 \mathbf{J}\ddot{\mathbf{q}}+\dot{\mathbf{J}}\dot{\mathbf{q}}=\mathbf{0},
\end{equation}
where $\mathbf{0}$ is a zero matrix with an appropriate dimension.

Combining Eqs.~\eqref{equ: dynamics} and \eqref{equ: holonomic constraint}, we compactly express the continuous-phase dynamics model as~\cite{gao2019global}:
\vspace{-0.2in}
\begin{equation}
\vspace{-0.2in}
\label{complete dynamics}
\mathbf{M}(\mathbf{q})\ddot{\mathbf{q}}+\bar{\mathbf{c}}(\mathbf{q},\dot{\mathbf{q}})=\mathbf{\bar B(\mathbf{q}) u},
\end{equation}
where the vector $\bar{\mathbf{c}}$ and matrix $\bar{\mathbf{B}}$ are defined as: $\bar{\mathbf{ c}}(\mathbf{q},\dot{\mathbf{q}}):=\mathbf{c}-\mathbf{J}^T(\mathbf{J} \mathbf{M}^{-1}\mathbf{J}^T)^{-1}(\mathbf{J} \mathbf{M}^{-1}\mathbf{c}-\dot{\mathbf{J}} \dot{\mathbf{q}})$ and $\mathbf{\bar B}(\mathbf{q}):=\mathbf{B}-\mathbf{J}^T(\mathbf{J} \mathbf{M}^{-1} \mathbf{J}^T)^{-1}\mathbf{J} \mathbf{M}^{-1}\mathbf{B}$.

\vspace{-0.2in}
\subsubsection{Switching surfaces}
When a robot's state reaches a switching surface, it exits the source domain and enters the targeted domain.
As displayed in Fig.~\ref{fig: multi_domain_walking1}, the three-domain walking involves three switching events, which are:
\begin{itemize}
    \item [(i)] Switching from FA to UA (``Support heel liftoff'');
    \item [(ii)] Switching from UA to OA (``Swing heel touchdown''); and
    \item [(iii)] Switching from OA to FA (``Leading toe touchdown'').
\end{itemize}
 
 The occurrence of these switching events is completely determined by the position and velocity of the robot's swing foot in the world frame as well as the ground-reaction force experienced by the support foot. 
 We use switching surfaces to describe the conditions under which a switching event occurs.

When the heel of the support foot takes off at the end of the FA phase, the robot enters the UA domain (Fig.~\ref{fig: multi_domain_walking1}).
This support heel liftoff condition can be described using the vertical ground-reaction force applied at the support heel, denoted as $F_{c,z}: \mathcal{T Q} \times {U} \rightarrow \mathbb{R}$.
We use $S_{F \rightarrow U}$ to denote the switching surface connecting an FA domain and its subsequent UA domain, and express it as:
\vspace{-0.2in}
\begin{equation*}
\vspace{-0.2in}
\label{switching1}
\begin{aligned}
	S_{F \rightarrow U}:=
 \{(\mathbf{q},\dot{\mathbf{q}},\mathbf{u})\in \mathcal{T Q} \times {U} : F_{c,z}(\mathbf{q},\dot{\mathbf{q}},\mathbf{u})&=0 \}.
\end{aligned}
\end{equation*} 

The UA$\rightarrow$OA switching occurs when the swing foot's heel lands on the ground (Fig.~\ref{fig: multi_domain_walking1}).
Accordingly, we express the switching surface that connects a UA domain and its subsequent OA domain, denoted as $S_{U \rightarrow O}$, as:
\vspace{-0.1in}
\begin{equation*}
\vspace{-0.1in}
	\label{switching2}
	S_{U \rightarrow O}(\mathbf{q},\dot{\mathbf{q}}):=\{(\mathbf{q},\dot{\mathbf{q}})\in \mathcal{T Q} : z_{swh}(\mathbf{q})=0,~\dot{z}_{swh}(\mathbf{q},\dot{\mathbf{q}})<0\},
\end{equation*} 
where $z_{swh}: \mathcal{Q}\rightarrow \mathbb{R}$ represents the height of the lowest point within the swing-foot heel above the ground.

As the leading toe touches the ground at the end of an OA phase, a new FA phase is activated (Fig.~\ref{fig: multi_domain_walking1}).
In this study, we assume that the leading toe landing and the trailing foot takeoff occur simultaneously at the end of an OA phase, which is reasonable because the trailing foot typically remains contact with the ground for a brief period (e.g., approximately 3$\%$ of a complete human gait cycle~\cite{zhao2017multi}) after the touchdown of the leading foot's toe.
The switching surface, $S_{O \rightarrow F}$, that connects an OA domain and its subsequent FA domain is then expressed as:
\vspace{-0.2in}
\begin{equation*}
\vspace{-0.1in}
\label{switching3}
S_{O \rightarrow F}(\mathbf{q},\dot{\mathbf{q}}):=\{(\mathbf{q},\dot{\mathbf{q}})\in \mathcal{T Q} : z_{swt}(\mathbf{q})=0,~\dot{z}_{swt}(\mathbf{q},\dot{\mathbf{q}})<0\},
\end{equation*}
where $z_{swt}: \mathcal{Q}\rightarrow \mathbb{R}$ represents the height of the swing-foot toe above the walking surface. 

\subsubsection{Discrete impact dynamics}

The complete walking cycle involves two foot-landing impacts; one impact occurs at the landing of the swing-foot heel (i.e., transition from UA to OA), and the other at the touchdown of the leading-foot toe between the OA and FA phases.
Note that the switching from FA to UA, characterized by the support heel liftoff, is a continuous process that does not induce any impacts.

We consider the case where the robot's feet and the ground are stiff enough to be considered as rigid, as summarized in the following assumptions~\cite{westervelt2007feedback,bhounsule2017discrete}:
\begin{itemize}
    \item [(A5)] The landing impact between the robot's foot and the ground is a contact between rigid bodies.
    \item[(A6)] The impact occurs instantaneously and lasts for an infinitesimal period of time.
\end{itemize}

Due to the impact between two rigid bodies (assumption (A5)), the robot's generalized velocity $\dot{\mathbf{q}}$ experiences a sudden jump upon a foot-landing impact.
Unlike velocity $\dot{\mathbf{q}}$, the configuration $\mathbf{q}$ remains continuous across an impact event as long as there is no coordinate swap of the two legs at any switching event.

Let $\dot{\mathbf{q}}^-$ and $\dot{\mathbf{q}}^+$ represent the values of $\dot{\mathbf{q}}$ just before and after an impact, respectively.
The impact dynamics can be described by the following nonlinear reset map~\cite{grizzle2001asymptotically}:
\vspace{-0.2in}
\begin{equation}
\vspace{-0.2in}
\label{reset-map-q}
\dot{\mathbf{q}}^+ = \boldsymbol{\Delta}_{\dot{q}} (\mathbf{q})\dot{\mathbf{q}}^-,	   
\end{equation}
where $\boldsymbol{\Delta}_{\dot{q}}: \mathcal{Q}\rightarrow \mathbb{R}^{(n+6) \times (n+6)}$ is a nonlinear matrix-valued function relating the pre-impact generalized velocity $\dot{\bq}^-$ to the post-impact value $\dot{\bq}^+$. 
The derivation of $\boldsymbol{\Delta}_{\dot{q}}$
is omitted and can be found in~\cite{westervelt2007feedback}.
Note that the dimension of $\boldsymbol{\Delta}_{\dot{q}}$ is invariant across the three domains since it characterizes the jumps of all floating-base generalized coordinates.

\section{CONTROLLER DESIGN FOR THREE-DOMAIN WALKING}
\label{Section-Control}

This section introduces the proposed GPT controller design based on the hybrid model of multi-domain bipedal robotic walking introduced in Section~\ref{Section-Dyanmics}. 
The resulting controller provably ensures the exponential error convergence for the directly regulated DOFs within each domain.
The sufficient conditions under which the proposed controller guarantees the stability for the overall hybrid system are provided in Section ~\ref{Section-Stability}.

\subsection{Desired Trajectory Encoding}

As the primary control objective is to provably drive the global-position tracking error to zero, one set of desired trajectories that the proposed controller aims to reliably track is the robot's desired global-position trajectories.
Since a bipedal humanoid robot typically has many more DOFs and actuators than the desired global-position trajectories, the controller could regulate additional variables of interest (e.g., swing-foot pose).

We use both time-based and state-based phase variables to encode these two sets of desired trajectories, as explained next.

\vspace{-0.2in}
\subsubsection{Time-based encoding variable}
\label{Sec: Control: Time-based}

We choose to use the global time variable $t$ to encode the desired global-position trajectories so that a robot's actual horizontal position trajectories in the world (i.e., $x_b$ and $y_b$) can be accurately controlled with precise timing, which is crucial for real-world tasks such as dynamic obstacle avoidance.

We use $x_d(t): \mathbb{R}^+ \rightarrow \mathbb{R}$ and $y_d(t): \mathbb{R}^+ \rightarrow \mathbb{R}$ to denote the desired global-position trajectories along the $x$- and $y$-axis of the world frame, respectively, and $\psi_d (t) : \mathbb{R}^+ \rightarrow \mathbb{R}$ is the desired heading direction.
We assume that the desired horizontal global-position trajectories $x_d(t)$ and $y_d(t)$ are supplied by a higher-layer planner, and the design of this planner is not the focus of this study.
Given $x_d(t)$ and $y_d(t)$, 
the desired heading direction $\psi_d (t) $ can be designed as a function of $x_d(t)$ and $y_d(t)$, which is $\psi_d (t) := \tan^{-1}(y_d/x_d)$.
Such a definition ensures that the robot is facing forward during walking.

We consider the following assumption on the regularity condition of $x_d(t)$ and $y_d(t)$:
\begin{itemize}
    \item [(A7)] The desired global-position trajectories $x_d(t)$ and $y_d(t)$ are planned as continuously differentiable on $t \in \mathbb{R}^+$ with the norm of $\dot{x}_d(t)$ and $\dot{y}_d(t)$ bounded above by a constant number; that is, there exists a positive constant $L_d$ such that
    \vspace{-0.2 in}
    \begin{equation}
    \vspace{-0.2 in}
        \| \dot{x}_d(t) \|,~\| \dot{y}_d(t)\|  \leq L_d
    \end{equation}
    for any $t \in \mathbb{R}^+$.
\end{itemize}
Under assumption (A7), the time functions $x_d(t)$ and $y_d(t)$ are Lipschitz continuous on $t \in \mathbb{R}^+$~\cite{khalil1996noninear}, which we utilize in the proposed stability analysis.

\vspace{-0.2in}
\subsubsection{State-based encoding variable}

As robotic walking inherently exhibits a cyclic movement pattern in the robot's configuration space, it is natural to encode the desired motion trajectories of the robot with a phase variable that represents the walking progress within a cycle.

To encode the desired trajectories other than the desired global-position trajectories, we choose to use a state-based phase variable, denoted $\theta(\mathbf{q}): \mathcal{Q} \rightarrow \mathbb{R}$, that represents the total horizontal distance traveled within a walking step.
Accordingly, the phase variable $\theta(\mathbf{q})$ increases monotonically within each walking step during straight-line or curved-path walking, which ensures a unique mapping from $\theta(\mathbf{q})$ to the encoded desired trajectories.
In contrast, in our previous work~\cite{gu2018straight,gu2022global}, the phase variable is chosen as the walking distance projected along a single direction on the ground, which may not ensure such a unique mapping during curved-path walking.

Since the phase variable $\theta(\mathbf{q})$ is essentially the length of a 2-D curve that represents the horizontal projection of the 3-D walking path on the ground,
we can use the actual horizontal velocities ($\dot{x}_b$ and $\dot{y}_b$) of the robot's base to express $\theta(\mathbf{q})$ as:
\vspace{-0.2in}
\begin{equation}
\vspace{-0.2in}
\label{equ:phase variable 1}
    \theta(\mathbf{q}(t)) = \int_{t_0}^{t} \sqrt{\dot{x}_b^2(t)+\dot{y}_b^2(t)}dt,
\end{equation}
where $t_0 \in \mathbb{R}^+$ represents the actual initial time instant of the given walking step and $t$ is the current time.

The normalized phase variable, which represents the percentage completion of a walking step, is given by: 
\vspace{-0.2in}
\begin{equation}
\vspace{-0.2in}
\label{equ:normalized phase variable}
s(\theta):=\frac{\theta}{\theta_{max}},
\end{equation}
where the real scalar parameter $\theta_{max}$ represents the maximum value of the phase variable (i.e., the planned total distance to be traveled within a walking step).
At the beginning of each step, the normalized phase variable takes a value of 0, while at the end of the step, it equals 1.

\vspace{-0.2in}
\subsection{Output Function Design}

An output function is a function that represents the difference between a control variable and its desired trajectory, which is essentially the trajectory tracking error.
The proposed controller aims to drive the output function to zero for the overall hybrid walking process.

Due to the distinct robot dynamics among different domains, we design different output functions (including the control variables and desired trajectories) for different domains.

\vspace{-0.2in}
\subsubsection{FA domain}

We use $\mathbf{h}_c^F(\mathbf{q}):\mathcal{Q} \rightarrow \mathbb{R}^{n}$ 
to denote the vector of $n$ control variables that are directly commanded within the FA domain.
Without loss of generality, we use the OP3 robot shown in Fig.~\ref{fig: op3 joints illustration} as an example to explain a common choice of control variables within the FA domain.

The OP3 robot has twenty directly actuated joints (i.e., $n=n_a=20$) including eight upper body joints.
Also, using $n_{up}$ to denote the number of upper body joints, we have $n_{up}=8$.

We choose the twenty control variables as follows: 
\begin{itemize}
    \item [(i)] The robot's global-position and orientation represented by the 6-D absolute base pose (i.e., position $\mathbf{p}_b$ and orientation $\boldsymbol{\gamma}_b$) w.r.t. the world frame;
    \item [(ii)] The position and orientation of the swing foot w.r.t the vehicle frame, respectively denoted as $\mathbf{p}_{sw}(\mathbf{q}): \mathcal{Q}\rightarrow\mathbb{R}^3$ and 
$\boldsymbol{\gamma_{sw}}(\mathbf{q}): \mathcal{Q}\rightarrow\mathbb{R}^3$; and 
    \item [(iii)] The angles of the $n_{up}$ upper body joints $\mathbf{q}_{up} \in \mathbb{R}^{n_{up}}$.
\end{itemize}

We choose to directly control the global-position of the robot to ensure that the robot's base follows the desired global-position trajectory.
The base orientation is also directly commanded to guarantee a steady trunk (e.g., for mounting cameras) and the desired heading direction. 
The swing foot pose is regulated to ensure an appropriate foot posture at the landing event, and
the upper body joints are controlled to avoid any unexpected arm motions that may affect the overall walking performance.

The stack of control variables $\mathbf{h}_c^F(\mathbf{q})$ are expressed as:

\begin{equation}
\small
\label{control variable full}
\mathbf{h}_c^F(\mathbf{q})
=
\begin{bmatrix}
    x_b\\
    y_b\\
    \psi_b\\
    z_b\\
    \phi_b\\
    \theta_b\\
    \mathbf{p}_{sw}\\
    \boldsymbol{\gamma}_{sw}\\
    \mathbf{q}_{up}
\end{bmatrix}.
\end{equation}

We use $\mathbf{h}_d^F(t,s): \mathbb{R}^+ \times [0,1] \rightarrow \mathbb{R}^{n}$ to denote the desired trajectories for the control variables $\mathbf{h}_c^F(\mathbf{q})$ within the FA domain.
These trajectories are encoded by the global time $t$ and the normalized state-based phase variable $s(\theta)$ as follows:
(i) the desired trajectories of the base position variables $x_b$ and $y_b$ and the base yaw angle $\psi_b$ are encoded by the global time $t$, while (ii) those of the other $(n-3)$ control variables, including the base height $z_b$, base roll angle $\phi_b$, base pitch angle $\theta_b$, swing-foot pose $\mathbf{p}_{sw}$
and $\boldsymbol{\gamma}_{sw}$, and upper joint angle $\mathbf{q}_{up}$, are encoded by the normalized phase variable $s(\theta)$.

The desired trajectory $\mathbf{h}_d^F(t,s)$ is expressed as: 
\vspace{-0.2in}
 \begin{equation}
 \vspace{-0.2in}
	\label{equ: had FD}
	\mathbf{h}_d^F(t,s)
	=
	\begin{bmatrix}
    x_d(t)\\
     y_d(t)\\
     \psi_d(t)\\ 
	\boldsymbol{\phi}^F(s)
	\end{bmatrix},
\end{equation}
where $x_d(t)$, $y_d(t)$, and $\psi_d (t) $ are defined in Section~\ref{Sec: Control: Time-based},
and the function $\boldsymbol{\phi}^F(s): [0,1] \rightarrow \mathbb{R}^{n-3} $ represents the desired trajectories of the control variables $z_b$, $\phi_b$, $\theta_b$, $\psi_b$, $\mathbf{p}_{sw}$, $\boldsymbol{\gamma}_{sw}$, and $\mathbf{q}_{up}$.

We use B\'ezier polynomials to parameterize the desired function $\boldsymbol{\phi}^F(s)$ because (i) they do not demonstrate overly large oscillations with relatively small parameter variations and (ii) their values at the initial and final instants within a continuous phase can compactly describe the values of control variables at those time instants~\cite{westervelt2007feedback}.

The desired function $\boldsymbol{\phi}^F(s)$ is given by:
\vspace{-0.2in}
\begin{equation} 
\label{Bezier polynomial}
\boldsymbol{\phi}^F_j(s):=\sum_{k=0}^{M} \alpha_{k,j}^F \frac{M!}{k!(M-k)!}s^k(1-s)^{M-k},
\vspace{-0.2in}
\end{equation}
where $\alpha_{k,j}^F \in \mathbb{R}$ ($k \in \{ 0,1,...,M \}$ and $j \in \{ 1,2,...,n-3 \}$) is the coefficient of the B\'ezier polynomials that are to be optimized (Section~\ref{Section-Simulation}), and $M$ is the order of the B\'ezier polynomials.

The output function during an FA phase is defined as:
\vspace{-0.2in}
\begin{equation}   
\vspace{-0.2in}
\mathbf{h}^F(t,\mathbf{q}):=\mathbf{h}_c^F(\mathbf{q})-\mathbf{h}_d^F(t,s).
        \label{equ: hd FD}
\end{equation}

\subsubsection{UA domain}
As explained in Section~\ref{Sec: walking domain}, a robot has $(n+1)$ DOF within the UA domain but only $n_a$ actuators.
Thus, only $n_a$ (i.e, $n$) variables can be directly commanded within the UA domain.

We opt to control individual joint angles within the UA domain to mimic human-like walking.
By ``locking'' the joint angles, the robot can perform a controlled falling about the support toe, similar to human walking.

Thus, the control variable $\mathbf{h}_{c}^U(\mathbf{q}): \mathcal{Q} \rightarrow \mathbb{R}^n$ is:
\begin{equation}
\begin{split} 
\label{control variable under}
\mathbf{h}_{c}^U(\mathbf{q})
=
\begin{bmatrix}
	 q_1\\
  q_2\\
  q_3\\
  ...\\
  q_n
\end{bmatrix}.
\end{split}
\end{equation}

Let 
$\mathbf{h}_{d}^U(s): [0,1] \rightarrow \mathbb{R}^{n}$
denote the desired joint position trajectories within the UA domain.
These desired trajectories $\mathbf{h}_{d}^U(s)$ are parameterized using B\'ezier polynomials $\boldsymbol{\phi}^U(s): [0,1] \rightarrow \mathbb{R}^{n}$; that is, $\mathbf{h}_{d}^U=\boldsymbol{\phi}^U(s)$.
The function $\boldsymbol{\phi}^U(s)$ can be expressed similarly to $\boldsymbol{\phi}^F(s)$.

The associated output function is then given by:
\begin{equation}
\mathbf{h}^U(\mathbf{q}):=\mathbf{h}_{c}^U(\mathbf{q})-\mathbf{h}_{d}^U(s).
\label{equ: desired walking pattern under}
\end{equation}

\vspace{-0.2 in}
\subsubsection{OA domain}
Let $\mathbf{h}_{c}^O(\mathbf{q}): \mathcal{Q} \rightarrow \mathbb{R}^{n-4}$ denote the control variables within the OA domain.
Recall that the robot has $n_a$ actuators and ($n-4$) DOFs within the OA domain.

We choose the ($n-4$) control variables as:
\begin{itemize}
    \item [(i)] The robot's 6-D base pose w.r.t. the world frame;
    \item [(ii)] The angles of the $n_{up}$ upper body joints, $\mathbf{q}_{up}$; and
     \item[(iii)] The pitch angles of the trailing and leading feet, denoted as $\theta_t(\mathbf{q})$ and $\theta_l(\mathbf{q})$, respectively.
\end{itemize}

Similar to the FA domain, we choose to directly command the robot's $6$-D base pose within the OA domain to ensure satisfactory global-position tracking performance, as well as the upper body joints to avoid unexpected arm movements that could compromise the robot's balance.
Also, regulating the pitch angle of the leading foot helps ensure a flat-foot posture upon switching into the subsequent FA domain where the support foot remains flat on the ground.
Meanwhile, controlling the pitch angle of the trailing foot can prevent overly early or late foot-ground contact events.

Thus, the control variable $\mathbf{h}_{c}^O(\mathbf{q})$ is:
\vspace{-0.2 in}
\begin{equation}
\vspace{-0.2 in}
\begin{split} 
\label{control variable over}
\mathbf{h}_{c}^O(\mathbf{q})
=
\begin{bmatrix}
x_b\\
y_b\\
\psi_b\\
z_b\\
\phi_b\\
\theta_b\\
\theta_t\\
\theta_l\\
\mathbf{q}_{up}
\end{bmatrix}.
\end{split}
\end{equation}

The desired trajectory
$\mathbf{h}_d^O(t,s): \mathbb{R}^+ \times [0,1] \rightarrow \mathbb{R}^{n-4}$ within the OA domain is expressed as:
\vspace{-0.2 in}
\begin{equation} 
\vspace{-0.2 in}
\label{desired_trajectory}
\mathbf{h}_{d}^O(t,s):= 
\begin{bmatrix}
x_d(t)\\
y_d(t)\\ 
\psi_d(t)\\
\boldsymbol{\phi}^O(s) 
\end{bmatrix},
\end{equation}
where 
$\boldsymbol{\phi}^O(s) : [0,1] \rightarrow \mathbb{R}^{n-4} $ represents the desired trajectories of $z_b$, $\phi_b$, $\theta_b$, $\theta_t$, $\theta_l$, and $\mathbf{q}_{up}$, which, similar to $\boldsymbol{\phi}^F(s)$ and $\boldsymbol{\phi}^U(s)$, can be chosen as B\'ezier curves.

The tracking error $\mathbf{h}^O(t,\bq)$ is expressed as:
\vspace{-0.2 in}
\begin{equation}
\mathbf{h}^O(t,\bq):=\mathbf{h}_{c}^O(\mathbf{q})-\mathbf{h}_{d}^O(t,s).
\label{desired walking pattern Over}
\end{equation}

\subsection{Input-Output Linearizing Control}
\label{Section-Control-Linearization}

The output functions representing the trajectory tracking errors can be compactly expressed as:
\vspace{-0.2 in}
\begin{equation}
\vspace{-0.2 in}
\label{output function}
{\mathbf{y}_i}= \mathbf{h}^i(t,\mathbf{q}),
\end{equation}
where the subscript $i \in \{F, U, O \}$ indicates the domain.

Due to the nonlinearity of the robot dynamics and the time-varying nature of the desired trajectories, the dynamics of the output functions are nonlinear and time-varying.
To reduce the complexity of controller design, we use input-output linearization to convert the nonlinear, time-varying error dynamics into a linear time-invariant one.

Let $\bu_i$ ($i \in \{F, U, O \}$) denote the joint torque vector within the given domain.
We exploit the input-output linearizing control law~\cite{khalil1996noninear}
\vspace{-0.1 in}
\begin{equation} 
\label{equ: IO-PD}
\mathbf{u}_i=(\tfrac{\partial \mathbf{h}^i}{\partial \mathbf{q}} \mathbf{M}^{-1}\bar{\mathbf{B}})^{-1}[(\tfrac{\partial \mathbf{h}^i}{\partial \mathbf{q}} )\mathbf{M}^{-1}\bar{\mathbf{c}}+\mathbf{v}_i-\tfrac{\partial^2 \mathbf{h}^i}{\partial t^2}-\tfrac{\partial}{\partial \mathbf{q}}(\tfrac{\partial \mathbf{h}^i}{\partial \mathbf{q}}\dot{\mathbf{q}})\dot{\mathbf{q}}]
\end{equation} 
to linearize the continuous-phase output function dynamics (i.e., Eq.~\eqref{complete dynamics}) into $\ddot{\mathbf{y}}_i=\mathbf{v}_i$, where $\mathbf{v}_i$ is the control law of the linearized system.
Here, the matrix $\tfrac{\partial \mathbf{h}^i}{\partial \mathbf{q}} \mathbf{M}^{-1}\bar{\mathbf{B}}$ is invertible on $\mathcal{Q}$ because
(i) $\bM$ is invertible on $\mathcal{Q}$,
(ii) $\tfrac{\partial \mathbf{h}^i}{\partial \mathbf{q}}$ is full row rank on $\mathcal{Q}$ by design, 
and (iii) $\bar{\bB}$ is full column rank on $\mathcal{Q}$. 

It should be noted that $\mathbf{u}_i$ has different expressions in different domains, due to the variations in the control variables and desired trajectories.
For instance, as the output function is time-independent within the UA domain, the function $\frac{\partial^2 \mathbf{h}^U}{\partial t^2}$ in Eq.~\eqref{equ: IO-PD} is always a zero vector because the output function $\mathbf{h}^U$ is explicitly time-independent. 

We design $\mathbf{v}_i$ as a proportional-derivative (PD) controller
\vspace{-0.2 in}
\begin{equation}
\vspace{-0.2 in}
    \mathbf{v}_i= -\mathbf{K}_{p, i} \mathbf{y}_i-\mathbf{K}_{d, i} \dot{\mathbf{y}}_i,
    \label{equ: IO-PD2}
\end{equation}
where $\bK_{p,i}$ and $\bK_{d,i}$ are positive-definite diagonal matrices containing the proportional and derivative control gains, respectively.
It is important to note that the dimension of the gains $\mathbf{K}_{p, i}$ and $\mathbf{K}_{d, i}$ depends on that of the output function in each domain;
their dimension is $n\times n$ in FA and UA domains, and $(n-4)\times (n-4)$ in the OA domain.

\begin{figure}[t]
    \centering
    \includegraphics[width=1\linewidth]{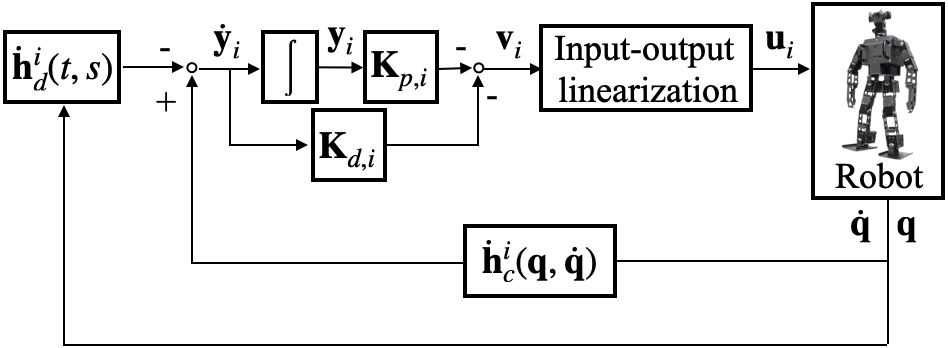}
    \vspace{-0.2in}
    \caption{Block diagram of the proposed global-position tracking control law within each domain. Here $i \in \{F,U,O \}$ indicates the domain type.}
    \label{fig:control_diagram}
    \vspace{-0.2in}
\end{figure}

We call the GPT control law in Eqs.~\eqref{equ: IO-PD} and \eqref{equ: IO-PD2} the ``IO-PD" controller in the rest of this paper, and the block diagram of the controller is shown in Fig.~\ref{fig:control_diagram}.

With the IO-PD control laws, the closed-loop output function dynamics within domain $i$ becomes linear:
\vspace{-0.2 in}
\begin{equation*}
\vspace{-0.2 in}
    \ddot{\mathbf{y}}_i=-\mathbf{K}_{d,i} \dot{\mathbf{y}}_i - \mathbf{K}_{p,i} \mathbf{y}_i.
\end{equation*}
Drawing upon the well-studied linear systems theory, we can ensure the exponential convergence of $\mathbf{y}_i$ to zero within each domain by properly choosing the values of the PD gain matrices ($\mathbf{K}_{p, i}$ and $\mathbf{K}_{d, i}$)~\cite{khalil1996noninear}.

\section{CLOSED-LOOP STABILITY ANALYSIS FOR THREE-DOMAIN WALKING}
\label{Section-Stability}

This section explains the proposed stability analysis of the closed-loop hybrid control system under the continuous IO-PD control law.

The continuous GPT law introduced in Section~\ref{Section-Control} with properly chosen PD gains achieves exponential stabilization of the output function state within each domain.
Nevertheless, the stability of the overall hybrid dynamical system is not automatically ensured for two main reasons.
First, within the UA domain, the utilization of the input-output linearization technique and the absence of actuators to directly control all the DOFs induce internal dynamics, which the control law cannot directly regulate~\cite{gu2017time,chan2018optimization}.
Second, the impact dynamics in Eq.~\eqref{reset-map-q} is uncontrolled due to the infinitesimal duration of an impact between rigid bodies (i.e., ground and swing foot).
As both internal dynamics and reset maps are highly nonlinear and time-varying, analyzing their effects on the overall system stability is not straightforward.

To ensure satisfactory tracking error convergence for the overall hybrid closed-loop system, we analyze the closed-loop stability via the construction of multiple Lyapunov functions~\cite{branicky1998multiple}.
The resulting sufficient stability conditions can be used to guide the parameter tuning of the proposed IO-PD law for ensuring system stability and satisfactory tracking.

\subsection{Hybrid Closed-Loop Dynamics}
This subsection introduces the hybrid closed-loop dynamics under the proposed IO-PD control law in Eqs.~\eqref{equ: IO-PD} and \eqref{equ: IO-PD2}, which serves as the basis of the proposed stability analysis.

\vspace{-0.2 in}
\subsubsection{State variables within different domains}
The state variables of the hybrid closed-loop system include the output function state ($\mathbf{y}_i$,$\dot{\mathbf{y}}_i$) ($i \in \{F,O,\xi\}$).
This choice of state variables allows our stability analysis to exploit the linear dynamics of the output function state within each domain, thus greatly reducing the complexity of the stability analysis for the hybrid, time-varying, nonlinear closed-loop system.

We use $ \mathbf{x}_F \in \mathbb{R}^{2n}$ and $ \mathbf{x}_O  \in \mathbb{R}^{2n-8}$ to respectively denote the state within the FA and OA domains, which are exactly the output function state:
\vspace{-0.2 in}
\begin{equation*}
\vspace{-0.2 in}
    \mathbf{x}_F := \begin{bmatrix} \mathbf{y}_F \\  \dot{\mathbf{y}}_F  \end{bmatrix}
    ~~~
    \text{and}
    ~~~
    \mathbf{x}_O := \begin{bmatrix} \mathbf{y}_O \\  \dot{\mathbf{y}}_O   \end{bmatrix}.
\end{equation*}

Within the UA domain, the output function state, denoted as $\bx_{\xi} \in \mathbb{R}^{2n-2}$, is expressed as:
\vspace{-0.2 in}
\begin{equation*}
\vspace{-0.2 in}
    \bx_{\xi} := \begin{bmatrix} \mathbf{y}_U \\  \dot{\mathbf{y}}_U   \end{bmatrix}.
\end{equation*}
Besides $\bx_{\xi}$, the complete state $\bx_{U}$ within the UA domain also include the uncontrolled state, denoted as $\bx_{\eta} \in \mathbb{R}^{2}$.
Since the stance-foot pitch angle $\theta_{st}(\bq)$ is not directly controlled within the UA domain, we define $\bx_{\eta}$ as: 
\vspace{-0.2 in}
\begin{equation*}
    \vspace{-0.2 in}
    \bx_{\eta} := 
    \begin{bmatrix} 
    \theta_{st}
    \\  \dot{\theta}_{st} 
    \end{bmatrix}.
\end{equation*}

Thus, the complete state within the UA domain is:
    \vspace{-0.2 in}
\begin{equation}
    \vspace{-0.2 in}
    \bx_U := 
    \begin{bmatrix}
        \bx_ {\xi} \\
        \bx_ {\eta}
    \end{bmatrix}.
\end{equation}

\subsubsection{Closed-loop error dynamics}
The hybrid closed-loop error dynamics associated with the FA and OA domains share the following similar form:
    \vspace{-0.2 in}
\begin{equation}
    \vspace{-0.2 in}
\begin{aligned}
    &
    \begin{cases}
    \dot{\mathbf{x}}_F = \mathbf{A}_F \mathbf{x}_F 
        &\text{if}~ 
                (t,\mathbf{x}_F^-) \notin  S_{\small F \rightarrow U}  \\
    \mathbf{x}_U^+ = \bm \Delta_{F \rightarrow U } (t, \mathbf{x}_F^-)  &\text{if}~ 
                (t,\mathbf{x}_F^-) \in  S_{F \rightarrow U}  \\
    \end{cases}
    \\
    &
    \begin{cases}
    \dot{\mathbf{x}}_O = \mathbf{A}_O \mathbf{x}_O 
        &\text{if}~ 
                (t,\mathbf{x}_O^-) \notin  S_{\small O \rightarrow F} \\
    \mathbf{x}_F^+ = \bm \Delta_{O \rightarrow F } (t, \mathbf{x}_O^-)  &\text{if}~ 
                (t,\mathbf{x}_O^-) \in  S_{O \rightarrow F}  \\
    \end{cases}
    \end{aligned}
    \label{Eq: closed-loop error dynamics}
\end{equation}
with
    \vspace{-0.2 in}
\begin{equation}
    \begin{aligned}
        \mathbf{A}_F:=\begin{bmatrix} 
	\mathbf{0} & \mathbf{I} \\
	-\mathbf{K}_{p,F} & -\mathbf{K}_{d,F} 
	\end{bmatrix}
 ~\text{and}~
 \mathbf{A}_O:=\begin{bmatrix} 
	\mathbf{0} & \mathbf{I} \\
	-\mathbf{K}_{p,O} & -\mathbf{K}_{d,O} 
	\end{bmatrix},
    \end{aligned}
    \label{Eq: closed-loop error dynamics2}
\end{equation}
where $ \mathbf{I}$ is an identity matrix with an appropriate dimension,
and $\bm \Delta_{F \rightarrow U}: \mathbb{R}^+ \times \mathbb{R}^{2n} \rightarrow \mathbb{R}^{2n+2}$
and
$\bm \Delta_{O \rightarrow F} : \mathbb{R}^+ \times \mathbb{R}^{2n-8} \rightarrow \mathbb{R}^{2n}$ are respectively the reset maps of the state vectors $\mathbf{x}_F$ and $\mathbf{x}_O$.
The expressions of $\bm \Delta_{F \rightarrow U}$ and $\bm \Delta_{O \rightarrow F}$ are omitted for space consideration and can be directly obtained by combining the expressions of the reset map $\boldsymbol{\Delta}_{\dq}$ of the generalized coordinates in Eq.~\eqref{reset-map-q} and the output functions $\by_F$, $\by_O$, and $\by_U$.

The closed-loop error dynamics associated with the continuous UA phase and the subsequent UA$\rightarrow$OA impact map can be expressed as:
    \vspace{-0.2 in}
\begin{equation}
\label{eq: xu dynamics}
    \begin{cases}
    \begin{cases}
            &\dot{\bx}_{\xi} = \mathbf{A}_{\xi} \bx_{\xi}  \\
            &\dot{\bx}_{\eta} = \mathbf{f}_{\eta} (t,\bx_{\eta}, \bx_{\xi})
    \end{cases}
        &\text{if}~ 
                (t,\mathbf{x}_U^-) \notin  S_{\small U \rightarrow O}  \\
    \bx_{O}^+ = \bm \Delta_{U \rightarrow O } (t,\bx_{\xi}^-,\bx_{\eta}^-) 
        &\text{if}~ 
                (t,\mathbf{x}_U^-) \in  S_{U \rightarrow O}  \\
    \end{cases}
    \\
\end{equation}
where
\begin{equation}
 \mathbf{A}_{\xi}:=\begin{bmatrix} 
	\mathbf{0} & \mathbf{I} \\
	-\mathbf{K}_{p,U} & -\mathbf{K}_{d,U} 
	\end{bmatrix}.
\end{equation}
The expression of $\mathbf{f}_{\eta}$ in Eq.~\eqref{eq: xu dynamics} can be directly derived using the continuous-phase dynamics equation of the generalized coordinates and the expression of the output function $\by_U$. 
Similar to $\bm \Delta_{F \rightarrow U}$ and $\bm \Delta_{O \rightarrow F }$, we can readily obtain the expression of the reset map $\bm \Delta_{U \rightarrow O}: \mathbb{R}^+ \times \mathbb{R}^{2n+2} \rightarrow \mathbb{R}^{2n-8}$ based on the reset map in Eq.~\eqref{reset-map-q} and the expression of $\by_U$ and $\by_O$.

\subsection{Multiple Lyapunov-Like Functions}

The proposed stability analysis via the construction of multiple Lyapunov functions begins with the design of the Lyapunov-like functions.
We use $V_{F}(\mathbf{x}_F)$, $V_{U}(\mathbf{x}_U)$, and $V_{O}(\mathbf{x}_O)$ to respectively denote the Lyapunov-like functions within the FA, UA, and OA domains, and introduce their mathematical expressions next.

\vspace{-0.2 in}
\subsubsection{FA and OA domains}

As the closed-loop error dynamics within the continuous FA and OA phases are linear and time-invariant, we can construct the Lyapunov-like functions $V_F (\bx_F)$ and $V_O (\bx_O)$ as~\cite{khalil1996nonlinear}:
\vspace{-0.2 in}
\begin{equation*}
\vspace{-0.2 in}
    V_F(\bx_F)=\bx_F^T \bP_F \bx_F
~\text{and}~
V_O(\bx_O)=\bx_O^T \bP_O \bx_O 
\end{equation*}
with $\bP_i$ ($i \in \{ F,O\}$) the solution to the Lyapunov equation
\vspace{-0.2 in}
\begin{equation*}
\vspace{-0.2 in}
    \bP_i \bA_i + \bA_i^T \bP_i = -\bQ_i,
\end{equation*}
where $\bQ_i$ is any symmetric positive-definite matrix with a proper dimension.

\subsubsection{UA domain}

As the input-output linearization technique is utilized and not all DOFs within the UA domain can be directly controlled, internal dynamics exist that cannot be directly controlled~\cite{gu2017time_thesis}. 
We design the Lyapunov-like function $V_U$ for the UA domain as:
\vspace{-0.2 in}
\begin{equation}
\vspace{-0.2 in}
    V_U = V_{\xi} (\mathbf{x}_{\xi}) + \beta \| \mathbf{x}_{\eta}  \|^2,
    \label{Vu-def}
\end{equation}
where $V_{\xi} (\mathbf{x}_{\xi})$ is a positive-definite function and $\beta$ is a positive constant to be designed.

As the dynamics of the output function state $\bx_{\xi}$ are linear and time-invariant, the construction of $V_{\xi} (\mathbf{x}_{\xi})$ is similar to that of $V_F$ and $V_O$:
\vspace{-0.2 in}
\begin{equation*}
\vspace{-0.2 in}
    V_{\xi}(\mathbf{x}_{\xi})= {\mathbf{x}_{\xi}}^T \bP_{\xi} {\mathbf{x}_{\xi}},
\end{equation*}
where $\bP_{\xi}$ is the solution to the Lyapunov equation
\vspace{-0.2 in}
\begin{equation*}
\vspace{-0.2 in}
    \bP_{\xi} \bA_{\xi} + \bA_{\xi}^T \bP_{\xi} = -\bQ_{\xi}
\end{equation*}
with $\bQ_{\xi}$ any symmetric positive-definite matrix with an appropriate dimension.

\subsection{Definition of Switching Instants}

\label{subsec: switching instants}

In the following stability analysis, the three domains of the $k^{th}$ ($k \in \{ 1,2,... \}$) walking step are, without loss of generality, ordered as:
\vspace{-0.2 in}
\begin{equation*}
\vspace{-0.2 in}
    FA \rightarrow UA \rightarrow OA.
\end{equation*}
For the $k^{th}$ walking step, we respectively denote the {\it actual} values of the initial time instant of the FA phase,
the $FA \rightarrow UA$ switching instant, the $UA \rightarrow OA$ switching instant, and the final time instant of the OA phase as: 
\vspace{-0.2 in}
\begin{equation*}
\vspace{-0.2 in}
    T_{3k-3},~T_{3k-2},~T_{3k-1},~\text{and}~T_{3k}.
\end{equation*}

The corresponding {\it desired} switching instants are denoted as:
\vspace{-0.2 in}
\begin{equation*}
\vspace{-0.2 in}
    \tau_{3k-3},~\tau_{3k-2},~\tau_{3k-1},~\text{and}~\tau_{3k}.
\end{equation*}

Using these notations, the $k^{th}$ actual complete gait cycle on $t\in (T_{3k-3},T_{3k})$ comprises:
\begin{itemize}
    \item [(i)] Continuous FA phase on $t \in (T_{3k-3},T_{3k-2})$;
    \item [(ii)] FA$\rightarrow$UA switching at $t = T_{3k-2}^-$;
    \item [(iii)] Continuous UA phase on $t \in (T_{3k-2},T_{3k-1})$;
    \item [(iv)] UA$\rightarrow$OA switching at $t = T_{3k-1}^-$;
    \item [(v)] Continuous OA phase on $t \in (T_{3k-1},T_{3k})$; and
    \item [(vi)] OA$\rightarrow$FA switching at $t = T_{3k}^-$.
\end{itemize}

For brevity in notation in the following analysis,
the values of any (scalar or vector) variable $\star$ at $t=T_{3k-j}^-$ and $t=T_{3k-j}^+$, i.e., 
\vspace{-0.2 in}
\begin{equation*}
    \star(T_{3k-j}^-)
    ~\mbox{and}~
    \star(T_{3k-j}^+),
    \vspace{-0.2 in}
\end{equation*}
are respectively denoted as:
\vspace{-0.2 in}
\begin{equation*}
   \star|^-_{3k-j}
    ~\mbox{and}~
   \star|^+_{3k-j}
   \vspace{-0.2 in}
\end{equation*}
for any $k \in \{ 1,2,... \}$ and $j \in \{0,1,2,3\}$.

\subsection{Continuous-Phase Convergence and Boundedness of Lyapunov-Like Functions}

As the output function state $\bx_i$ ($i \in \{F,O,\xi \}$) is directly controlled, we can readily analyze the convergence of the output functions (and their associated Lyapunov-like functions, $V_F$, $V_O$, and $V_{\xi}$) within each domain based on the well-studied linear systems theory~\cite{khalil1996noninear}.

\begin{prop}\textup{(\textbf{Continuous-phase output function convergence within each domain})}
\label{prop: 1}
Consider the IO-PD control law in Eq.~\eqref{equ: IO-PD}, assumptions (A1)-(A7), and the following condition:
\begin{itemize}
    \item [(B1)]  The PD gains are selected such that $\mathbf{A}_F$, $\mathbf{A}_O$, and $\mathbf{A}_{\xi}$ are Hurwitz.
 \end{itemize}
Then, there exist positive constants $r_i$, $c_{1i}$, $c_{2i}$, and $c_{3i}$ ($i \in \{ F,O,\xi \}$) such that the Lyapunov-like functions $V_F$, $V_O$, and $V_{\xi}$ satisfy the following inequalities
\vspace{-0.2 in}
\begin{equation}
\vspace{-0.2 in}
    c_{1i}  \|   \mathbf{x}_i    \|^2   
    \leq 
    V_i (  \mathbf{x}_i  ) 
    \leq 	
    c_{2i}  \|   \mathbf{x}_i    \|^2  
    ~~~~
    \text{and}
    ~~~~
    \dot{V}_{i}  \leq 	- c_{3i} {V}_{i}
    \label{Lyap-F-O}
\end{equation}
within their respective domains for any 
\vspace{-0.2 in}
\begin{equation*}
    \vspace{-0.2 in}
    \mathbf{x}_i \in B_{r_i}(\mathbf{0}) := 
\{\mathbf{x}_i:  \| \mathbf{x}_i \| \leq r_i  \}, 
\end{equation*}
where $\bzero$ is a zero vector with an appropriate dimension.

Moreover, Eq.~\eqref{Lyap-F-O} yields 
\vspace{-0.2 in}
\begin{equation}
\vspace{-0.2 in}
     V_F|^-_{3k-2} \leq e^{-c_{3F} ( T_{3k-2} - T_{3k-3} )} V_F|^+_{3k-3},
     \label{eq: VF-cont}
\end{equation}
\vspace{-0.2 in}
\begin{equation}
\vspace{-0.2 in}
     V_O|^-_{3k} \leq e^{-c_{3O} ( T_{3k} - T_{3k-1} )} V_O|^+_{3k-1},
      \label{eq: VO-cont}
\end{equation} 
and
\vspace{-0.2 in}
\begin{equation}
\vspace{-0.2 in}
     V_{\xi}|^-_{3k-1} 
     \leq 
     e^{-c_{3\xi} ( T_{3k-1} - T_{3k-2} )} V_{\xi}|^+_{3k-2},
      \label{eq: Vxi-cont}
\end{equation}
which describe the exponential continuous-phase convergence of $V_F$, $V_O$, and $V_{\xi}$ within their respective domains.
\end{prop}

The proof of Proposition \ref{prop: 1} is omitted as Proposition \ref{prop: 1} is a direct adaptation of the Lyapunov stability theorems from~\cite{khalil1996noninear}.
Note that the explicit relationship between the PD gains and the continuous-phase convergence rates $c_{3F}$, $c_{3O}$, and $c_{3\xi}$ can be readily obtained based on Remark 6 of our previous work~\cite{gu2022global}.

Due to the existence of the uncontrolled internal state, the Lyapunov-like function $V_U$ does not necessarily converge within the UA domain despite the exponential continuous-phase convergence of $V_{\xi}$ guaranteed by the proposed IO-PD control law that satisfies condition (B1).
Still, we can prove that within the UA domain of any $k^{th}$ walking step, the value of the Lyapunov-like function $V_{U}$ just before switching out of the domain, i.e., $V_U |^-_{3k-1}$, is bounded above by a positive-definite function of the ``switching-in'' value of $V_{U}$, i.e., $V_U |^+_{3k-2}$, as summarized in Proposition 2.

\begin{prop}\textup{(\textbf{Boundedness of Lyapunov-like function within UA domain})}
\label{prop: 2}
{\it 
Consider the IO-PD control law in Eq.~\eqref{equ: IO-PD} and all conditions in Proposition~\ref{prop: 1}. 
There exists a positive real number $r_{U1}$ and a positive-definite function $w_u(\cdot)$ such that}
\vspace{-0.1 in}
\begin{equation*}
\vspace{-0.1 in}
    V_U |^-_{3k-1}  \leq w_u(V_U |^+_{3k-2}  )
\end{equation*}
{\it holds for any $k \in \{1,2,...\}$ and $\mathbf{x}_{U} \in B_{r_{U1}} (\mathbf{0})$.}
\end{prop}

\noindent \textbf{Rationale of proof}:
The proof of Proposition~\ref{prop: 2} is given in Appendix~\ref{Appendix: Prop2}.
The boundedness of the Lyapunov-like function $V_U(\bx_U)$ at $t=T^-_{3k-1}$ is proved based on the definition of $V_U(\bx_U)$ given in Eq.~\eqref{Vu-def} and the boundedness of $\Big\| \bx_U |^-_{3k-1} \Big\|$.
Recall 
$\bx_U := 
    \begin{bmatrix}
        \bx_ {\xi}^T&
        \bx_ {\eta}^T
    \end{bmatrix}^T.$
We establish the needed bound on $\Big\| \bx_U |^-_{3k-1} \Big\|$ through the bounds on 
$\Big\| \bx_{\xi} |^-_{3k-1} \Big\|$
and
$\Big\| \bx_{\eta} |^-_{3k-1} \Big\|$,
which are respectively obtained based on the bounds of their continuous-phase dynamics of $\bx_ {\xi}$ and $\bx_ {\eta}$ and the integration of those bounds within the given continuous UA phase.
$\hfill
\blacksquare$

\vspace{-0.2 in}
\subsection{Boundedness of Lyapunov-Like Functions across Jumps}

\begin{prop}\textup{(\textbf{Boundedness across jumps})}
\label{prop: 3}
Consider the IO-PD control law in Eq.~\eqref{equ: IO-PD}, all conditions in Proposition~\ref{prop: 1}, and the following two additional conditions:
\begin{itemize}
    \item [(B2)] The desired trajectories $\bh_d^i$ ($i \in \{F, U, O \}$) are planned to respect the impact dynamics with a small, constant offset ${\gamma}_{\Delta}$; that is,
\end{itemize}
\vspace{-0.2 in}
\begin{align}
\vspace{-0.2 in}
& \|  \bm \Delta_{F \rightarrow U} (\tau_{3k-2},\mathbf{0}) \| \leq \gamma_\Delta,
\label{reset_f2u}
\\
&  \| \bm \Delta_{U \rightarrow O} (\tau_{3k-1},\mathbf{0}) \| \leq \gamma_\Delta  ,~\text{and}~
\\
&  \| \bm \Delta_{O \rightarrow F} (\tau_{3k},\mathbf{0}) \| \leq \gamma_\Delta.
\end{align}
\begin{itemize}
    \item [(B3)] The PD gains are chosen to ensure a sufficiently high convergence rate (i.e., $c_{3F}$, $c_{3O}$, and $c_{3\xi}$ in Eqs.~\eqref{eq: VF-cont}-\eqref{eq: Vxi-cont}) of $V_F$, $V_O$, and $V_{\xi}$.
\end{itemize}
Then, there exists a positive real number $r$ such that for any $k \in \{1,2,...\}$, $\mathbf{x}_i\in B_{r} (\mathbf{0})$, and $i \in \{ F,U,O\}$, the following inequalities
\begin{equation}
\begin{aligned}
&... \leq  V_F|_{3k}^+  \leq  V_F|_{3k-3}^+ \leq ... \leq V_F|_{3}^+ \leq  V_F|_{0}^+,
  \\
&... \leq V_U|_{3k+1}^+  \leq  V_U|_{3k-2}^+ \leq ... \leq V_U|_{4}^+ \leq V_U|_{1}^+,
\\
~\text{and}~&
  \\
&...  \leq  V_O|_{3k+2}^+  \leq  V_O|_{3k-1}^+ \leq ... \leq V_F|_{5}^+ \leq  V_F|_{2}^+
\end{aligned}
\label{eq: lyapunov boundedness}
\end{equation}
hold;
that is, the values of each Lyapunov-like function at their associated ``switching-in" instants form a nonincreasing sequence.
\end{prop}

\noindent \textbf{Rationale of proof}:
The proof of Proposition~\ref{prop: 3} is given in Appendix~\ref{Appendix: Prop3}.
The proof shows the derivation details for the first inequality in Eq.~\eqref{eq: lyapunov boundedness} (i.e., $ V_F|_{3k}^+  \leq  V_F|_{3k-3}^+$ for any $k \in \{ 1,2,... \}$), which can be readily extended to prove the other two inequalities.

The proposed proof begins the analysis of the time evolution of the three Lyapunov-like functions within a complete gait cycle from $t=T^+_{3k-1}$ to $t=T^+_{3k}$, which comprises three continuous phases and three switching events as listed in Section~\ref{subsec: switching instants}.

Based on the time evolution, the bounds on the Lyapunov-like functions $V_F$, $V_O$, and $V_U$ at the end of their respective continuous phases are given in Proposition~\ref{prop: 1} and \ref{prop: 2}, while their bounds at the beginning of those continuous phases are established through the analysis of the reset maps $\boldsymbol{\Delta}_{F \rightarrow U}$, $\boldsymbol{\Delta}_{U \rightarrow O}$, and $\boldsymbol{\Delta}_{O \rightarrow F}$.
Finally, we combine these bounds to prove $ V_F|_{3k}^+  \leq  V_F|_{3k-3}^+$.
$\hfill
\blacksquare$

The offset ${\gamma}_{\Delta}$ is introduced in condition (B2) for two primary reasons.
Firstly, since the system's actual state trajectories inherently possess the impact dynamics, the desired trajectories need to respect the impact dynamics sufficiently closely (i.e., ${\gamma}_{\Delta}$ is small enough) in order to avoid overly large errors after an impact~\cite{rijnen2019hybrid,rijnen2016hybrid}.
If the desired trajectories do not agree with the impact dynamics sufficiently closely, the tracking errors at the beginning of a continuous phase could be overly large even when the errors at the end of the previous continuous phase are small.
Such error expansion could induce aggressive control efforts at the beginning of a continuous phase, which could reduce energy efficiency and might even cause torque saturation.
Secondly, while it is necessary to enforce the desired trajectories to respect the impact dynamics (e.g., through motion planning), requiring the exact agreement with the highly nonlinear impact dynamics (i.e., ${\gamma}_{\Delta}=0$) could significantly increase the computationally burden of planning, which could be mitigated by allowing a small offset.

\subsection{Main Stability Theorem}

We derive the stability conditions for the hybrid error system in Eqs.~\eqref{Eq: closed-loop error dynamics} and \eqref{eq: xu dynamics} based on Propositions 1-3 and the general stability theory via the construction of multiple Lyapunov functions~\cite{branicky1998multiple}.

\begin{thm}(\textup{\textbf{Closed-loop stability conditions}})
\label{thm: 1}
Consider the IO-PD control law in Eq.~\eqref{equ: IO-PD}.
If all conditions in Proposition~\ref{prop: 3} are met, 
the origin of the hybrid closed-loop error system in Eqs.~\eqref{Eq: closed-loop error dynamics} and \eqref{eq: xu dynamics} is locally stable in the sense of Lyapunov.
\end{thm}

\noindent \textbf{Rationale of proof}:
The full proof of Theorem~\ref{thm: 1} is given in Appendix~\ref{Appendix: thm1}.
The key idea of the proof is to show that the closed-loop control system satisfies the general multiple-Lyapunov stability conditions given in ~\cite{branicky1998multiple} if all conditions in Proposition \ref{prop: 3} are met.
$\hfill
\blacksquare$

\section{EXTENSION FROM THREE-DOMAIN WALKING WITH FULL MOTOR ACTIVATION TO TWO-DOMAIN WALKING WITH INACTIVE ANKLE MOTORS}
\label{Section-Second MD walking}

This section explains the design of a GPT control law for a two-domain walking gait to further illustrate the proposed controller design method.
The controller is a direct extension of the proposed controller design for three-domain walking (with full motor activation).  
For brevity, this section focuses on describing the distinct aspects of the two-domain case compared to the three-domain case explained earlier.

We consider the case of two-domain walking where underactuation is caused due to intentional ankle motor deactivation instead of loss of full contact with the ground as in the case of three-domain walking.
Bipedal gait is sometimes intentionally designed as underactuated through motor deactivation at the support ankle~\cite{gong2020angular}, which could simplify the controller design.
Specifically, by switching off the support ankle motors, the controller can treat the support foot as part of the ground
and only handle a point foot-ground contact instead of a finite support polygon.

\begin{figure}[t]
\centering
\includegraphics[width=0.7\linewidth]{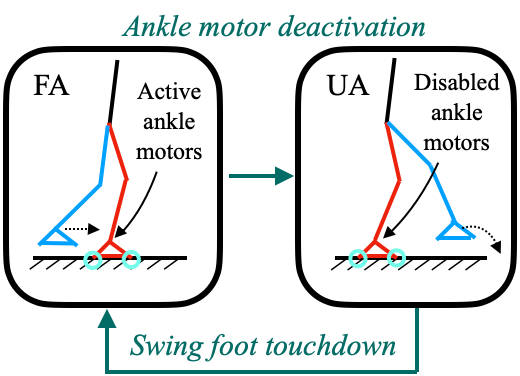}
	\caption{Illustration of a complete two-domain walking cycle.
   The green circles show the portions of the feet that touch the ground. The leg in red represents the support leg, while the leg in blue the swing leg. The movement of the swing foot is shown by the dashed arrow.
	}
	\label{fig: multi_domain 2}
 \vspace{-0.1 in}
\end{figure}

Figure~\ref{fig: multi_domain 2} illustrates a complete cycle of a two-domain walking gait, which comprises an FA and a UA domain, with the UA phase induced by intentional motor deactivation.
The FA and UA phases share the same foot-ground contact conditions; that is, the toe and heel of the support foot are in a static contact with the ground. 
Yet, within the UA domain, the ankle-roll and ankle-pitch joints of the support foot are disabled, leading to $\text{DOF}=n_a+2>n_a$ (i.e., underactuation). 

To differentiate from the case of three-domain walking, we add a ``$\dagger$'' superscript to the left of mathematical symbols when introducing the two-domain case.

\noindent \textbf{Hybrid robot dynamics:}
The continuous-time robot dynamics within the FA domain of two-domain walking have exactly the same expression as those of the three-domain dynamics in Eq.~\eqref{equ: dynamics}.
The robot dynamics within the UA domain are also the same as Eq.~\eqref{equ: dynamics} except for the input matrix $\mathbf{B}$ (due to the ankle motor deactivation).

The complete gait cycle contains one foot-landing impact event, which occurs as the robot's state leaves the UA domain and enters the FA domain.
The form of the associated impact map is similar to the impact map in Eq.~\eqref{reset-map-q} of the three-domain case.
For brevity, we omit the expression and derivation details of the impact map.

There are two switching events, F$\rightarrow$U and U$\rightarrow$F, within a complete gait cycle, which are respectively denoted as $^\dagger S_{F \rightarrow U}$ and $^\dagger S_{U \rightarrow F}$ and given by:
\begin{equation*}
\label{switching2}
\begin{aligned}
    ^\dagger S_{F \rightarrow U}&:=\{\mathbf{q}\in\mathcal{Q}:  \theta(\mathbf{q})>l_s \} ~\text{and}~
\\
^\dagger S_{U \rightarrow F}&:=\{(\mathbf{q},\dot{\mathbf{q}})\in  \mathcal{TQ} : z_{sw}(\mathbf{q})=0,\dot{z}_{sw}(\mathbf{q},\dot{\mathbf{q}})<0\},
\end{aligned}
\end{equation*}
where 
$\theta(\mathbf{q})$ is defined as in Eq.~\eqref{theta} and
the scalar positive variable $l_s$ represents the desired traveling distance of the robot's base within the FA phase.

\noindent \textbf{Local time-based phase variable:}
To allow the convenient adjustment of the intended period of motor deactivation, we introduce a new phase variable $\prescript{\dagger}{}{\theta}(t)$ for the UA phase representing the elapsed time within this phase:
$\prescript{\dagger}{}{\theta}(t) = t-T_{Uk}$,
where $T_{Uk}$ is the initial time instant of the $k^{th}$ UA phase.

The normalized phase variable is defined as:
$\prescript{\dagger}{}{s}( \prescript{\dagger}{}{\theta} ):= \frac{   \prescript{\dagger}{}{\theta}  }{\delta_{\tau_U}}$, where $\delta_{\tau_U}$ is the expected duration of the UA. $\delta_{\tau_U}$ can be assigned as a gait parameter that a motion planner adjusts for ensuring a reasonable duration of motor deactivation.

\noindent \textbf{Output functions:}
The output function design within the FA domain is the same as the three-domain case.

The control variables within FA, denoted as $\prescript{\dagger}{}{\mathbf{h}_c^F(\mathbf{q})}$, are chosen the same as the three-domain walking case in Eq.~\eqref{control variable full}. 
Then, we have
$\prescript{\dagger}{}{\mathbf{h}_c^F}(\mathbf{q})=\mathbf{h}_c^F(\mathbf{q})$.
Accordingly, the desired trajectories $\prescript{\dagger}{}{\mathbf{h}_d^F}(t, s)$ can be chosen the same as $\mathbf{h}_d^F(t, s)$, leading to the output function expressed as: $\prescript{\dagger}{}{\mathbf{h}^F}(t, s)=\prescript{\dagger}{}{\mathbf{h}_c^F}(\mathbf{q})-\prescript{\dagger}{}{\mathbf{h}_d^F}(t, s)$.

With two ankle (roll and pitch) motors disabled during the UA phase, the number of variables that can be directly controlled is reduced by two compared to the FA domain. 
Without loss of generality, 
We choose the control variables within the UA domain to be the same as the FA domain except that the base roll angle $\phi_b$ and base pitch angle $\theta_b$ are no longer controlled.

The control variables $^\dagger\mathbf{h}_c^U$ within the UA domain are then expressed as:
\vspace{-0.1 in}
\begin{equation}
\vspace{-0.1 in}
\begin{split} 
\label{equ: control variable UD}
^\dagger\mathbf{h}_c^U(\mathbf{q})
:=
\begin{bmatrix}
x_b
\\
y_b\\
\psi_b\\
z_b\\
\mathbf{p}_{sw}(\mathbf{q})\\
\boldsymbol{\gamma}_{sw}(\mathbf{q})\\
\end{bmatrix}.
\end{split}
\end{equation}
 
The desired trajectories $^\dagger\mathbf{h}_d^U$ are given by:
\vspace{-0.1 in}
\begin{equation}
\vspace{-0.1 in}
\label{equ: hd UD}
^\dagger\mathbf{h}_d^U(t,\prescript{\dagger}{}{s})
:=
\begin{bmatrix}
    x_d(t) \\
    y_d(t) \\
    \psi_d(t)\\
    \prescript{\dagger}{}{\boldsymbol{\phi}^U}(\prescript{\dagger}{}{s})
\end{bmatrix},
\end{equation}
where $\prescript{\dagger}{}{\boldsymbol{\phi}^U}(\prescript{\dagger}{}{s}): [0,1] \rightarrow \mathbb{R}^{{n_a}-5}$ represents the desired trajectories of $z_b$, $\mathbf{p}_{sw}$, and $\boldsymbol{\gamma}_{sw}$.

Then, we obtain the output function $^\dagger\mathbf{h}^U(t,\mathbf{q})$ as:
    \vspace{-0.2 in}
\begin{equation}
\vspace{-0.2 in}
^\dagger\mathbf{h}^U(t,\mathbf{q}):=~^\dagger\mathbf{h}_c^U(\mathbf{q})-~^\dagger\mathbf{h}_d^U(t,\prescript{\dagger}{}{s}).
\label{equ: hd FD}
\end{equation}

With the output function $^\dagger\mathbf{h}^i$ ($i \in \{F,A,U \}$) designed, we can use the same form of the IO-PD control law in Eqs.~\eqref{equ: IO-PD} and ~\eqref{equ: IO-QP} and the stability conditions in Theorem 1 to design the needed GPT controller for two-domain walking.

\section{SIMULATION}	
\label{Section-Simulation}
This section reports the simulation results to demonstrate the satisfactory global-position tracking performance of the proposed controller design.

\subsection{Comparative Controller: 
Input-Output Linearizing Control with Quadratic Programming}	

This subsection introduces the formulation of the proposed IO-PD controller as a quadratic program (QP) that handles the limited joint-torque capacities of real-world robots while ensuring a relatively accurate global-position tracking performance.
We refer to the resulting controller as the ``IO-QP'' controller in this paper.
Besides enforcing the actuator limits and providing tracking performance guarantees, another benefit of the QP formulation lies in its computational efficiency for real-time implementation. 

\vspace{-0.2 in}
\subsubsection{Constraints}

We incorporate the IO-PD controller in Eq.~\eqref{equ: IO-PD} as an equality constraint in the proposed IO-QP control law.
The proposed IO-QP also includes the torque limits as inequality constraints.
We use $\mathbf{u}_{max,i}$ and $\mathbf{u}_{min,i}$ ($i \in \{F,U,O \}$) to denote the upper and lower limits of the torque command $\mathbf{u}_i$ given in Eq.~\eqref{equ: IO-PD}.
Then, the linear inequality constraint that the control signal $\mathbf{u}_i$ should respect can be expressed as: 
$\mathbf{u}_{min,i} \le \mathbf{u}_i \le \mathbf{u}_{max,i}$.

To ensure the control command $\mathbf{u}_i$ respects the actuator limits,
we incorporate a slack variable $\boldsymbol{\delta_{QP}}\in\mathbb{R}^{n_a}$ in the equality constraint representing the IO-PD control law:
\vspace{-0.2 in}
\begin{equation}
\vspace{-0.2 in}
\label{equ:IOQP}
    \mathbf{u}_i = {\mathbf{N}}(\mathbf{q},\dot{\mathbf{q}}) +\boldsymbol{\delta}_{QP},
\end{equation}
where ${\mathbf{N}}=(\tfrac{\partial \mathbf{h}^i}{\partial \mathbf{q}} \mathbf{M}^{-1}\bar{\mathbf{B}})^{-1}[(\tfrac{\partial \mathbf{h}^i}{\partial \mathbf{q}} )\mathbf{M}^{-1}\bar{\mathbf{c}}+\mathbf{v}_i-\tfrac{\partial^2 \mathbf{h}^i}{\partial t^2}-\tfrac{\partial}{\partial \mathbf{q}}(\tfrac{\partial \mathbf{h}^i}{\partial \mathbf{q}}\dot{\mathbf{q}})\dot{\mathbf{q}}]$.
To avoid overly large deviation from the original control law in Eq.~\eqref{equ: IO-PD}, we include the slack variable in the cost function to minimize its norm as explained next.

\vspace{-0.2 in}
\subsubsection{Cost function}

The proposed cost function is the sum of two components.
One term is $\mathbf{u}_i^T \mathbf{u}_i$ and indicates the magnitude of the control command $\mathbf{u}_i$.
Minimizing this term helps guarantee the satisfaction of the torque limit and the energy efficiency of walking.

The other term indicates the weighted norm of the slack variable $\boldsymbol{\delta}_{QP}$, i.e., $p\boldsymbol{\delta_{QP}}^T\boldsymbol{\delta_{QP}}$, with the real positive scalar constant $p$ the slack penalty weight.
By including the slack penalty term in the cost function, the deviation of the control signal from the original IO-PD form, which is caused by the relaxation, can be minimized.  

\vspace{-0.2 in}
\subsubsection{QP formulation}

Summarizing the constraints and cost function introduced earlier, we arrive at a QP given by:
\vspace{-0.2 in}
\begin{equation}
\vspace{-0.2 in}
    \label{equ: IO-QP}
    \begin{aligned}
        \underset {\mathbf{u}_i,\boldsymbol{\delta}_{QP}}{\min } & \qquad \mathbf{u}_i^{T} \mathbf{u}_i +p~\boldsymbol{\delta}_{QP}^T\boldsymbol{\delta}_{QP}  
        \\
        \text {s.t.} & \qquad
        \mathbf{u}_i = {\mathbf{N}} +\boldsymbol{\delta}_{QP}
        \\
        & \qquad \mathbf{u}_i \ge \mathbf{u}_{min,i}
        \\
        & \qquad \mathbf{u}_i \le \mathbf{u}_{max,i} 
    \end{aligned}
\end{equation}

We present validation results for both IO-PD and IO-QP in the following to demonstrate their effectiveness and performance comparison.

\begin{table}[t]
\centering
\vspace{-0.05in}
\caption{\small{Mass distribution of the OP3 robot.}} 
\small
\begin{tabular}{ p{3.4cm}|p{1.6cm}|p{1.6cm} }
\hline
\hline
\centering
Body component & \centering  Mass (kg) & {\centering  Length (cm)} \\
\hline
\centering trunk & \centering 1.34 & ~~~~~~~~63\\
\centering left/right thigh & \centering 0.31& {\centering ~~~~~~~~11}\\
\centering left/right shank & \centering 0.22& ~~~~~~~~11\\
\centering left/right foot & \centering 0.07& ~~~~~~~~12\\
\centering left/right upper arm & \centering 0.19& ~~~~~~~~12\\
\centering left/right lower arm & \centering 0.04& ~~~~~~~~12\\
\centering head & \centering 0.15& ~~~~~~~N/A\\
\hline
\hline
\end{tabular}
\label{Table: robot info}
\end{table}

\subsection{Simulation Setup}

\label{Sec: Sim Setup}

\subsubsection{Robot model}
The robot used to validate the proposed control approach is an OP3 bipedal humanoid robot developed by ROBOTIS, Inc. (see Fig.\ref{fig: op3 joints illustration}). 
The OP3 robot is $50$ cm tall and weighs approximately $3.2$ kg.
It is equipped with $20$ active joints, as shown in Fig.~\ref{fig: op3 joints illustration}.
The mass distribution and geometric specifications of the robot are listed in Table~\ref{Table: robot info}.~To validate the proposed controller, we use the MATLAB ODE solver {\tt ODE45} to simulate the dynamics models of the OP3 robot for both three-domain walking (Section 2) and two-domain walking (Section 5).
The default tolerance settings of the {\tt ODE45} solver are used.

\vspace{-0.2 in}
\subsubsection{Desired global-position trajectories and walking patterns}

As mentioned earlier, this study assumes that the desired global-position trajectories are provided by a higher-layer planner.~To assess the effectiveness of the proposed controller, three different desired global-position (GP) trajectories are tested, including single-direction and varying-direction trajectories. These trajectories are specified in Table.~\ref{Table: GP}.

\begin{table}[t]
\centering
\vspace{-0.05in}
\caption{\small{Desired global-position trajectories.}} 
\small
\begin{tabular}{ p{0.7cm}|p{2.1cm}|p{2.2cm}|p{1.2cm}  }
\hline
\hline
\centering
Traj. index & \centering  $x_d(t)$~(cm) & \centering  $y_d(t)$~(cm)& Time interval~(s) \\
\hline
\centering (GP1) & \centering \centering$8 t$ & \centering$0$ & ~~$[0, +\infty)$\\
\hline
\centering (GP2) & \centering $19.1t$& \centering $5.9t$&~~$[0, +\infty)$\\
\hline
\multirow{4}{*}{(GP3)} & \centering $25 t$& \centering$0$& $[0, 3.13)$
\\
& \centering $3000 \sin(\tfrac{t-3.13}{80})+78.2$& \centering$3000 \cos(\tfrac{t-3.13}{80})-3000$& $[3.13, 4.25)$\\
& \centering $24(t-4.25)+120$& \centering $-7(t-4.25)-0.3$ & $[4.25, +\infty)$\\
\hline
\hline
\end{tabular}
\label{Table: GP}
\end{table}

The GPs include two straight-line global-position trajectories with distinct
heading directions, labeled as (GP1) and (GP2).
We set the velocities of (GP1) and (GP2) to be different to evaluate the performance of the controller under different walking speeds.
To assess the effectiveness of the proposed control law in tracking the desired global-position trajectories along a path with different walking directions, we also consider a walking trajectory (GP3) consisting of two straight-line segments connected via an arc.

\begin{table}[t]
\centering
\vspace{-0.05in}
\caption{\small{Initial tracking error norms for three cases.}
} 
\small
\begin{tabular}{ p{3.2cm}|p{1cm}|p{1cm}|p{1cm}  }
\hline
\hline
\centering Tracking error norm & Case A & \centering  Case B &  Case C
\\
\hline
\centering swing foot position (\% of step length)& \centering \multirow{2}{*}{27.5} & \centering\multirow{2}{*}{27.5}& \multirow{2}{*}{~~~~40} 
\\ 
\hline
\centering base orientation (deg.) & \centering 0 & \centering 17& ~~~~12  \\
\hline
\centering base position (\% of step length) & \centering \multirow{2}{*}{15} & \centering \multirow{2}{*}{15} &  \multirow{2}{*}{~~~~~8} \\
\hline
\hline
\end{tabular}
\label{Table: initial error}
\end{table}

\begin{figure}
    \centering    \includegraphics[width=1\linewidth]{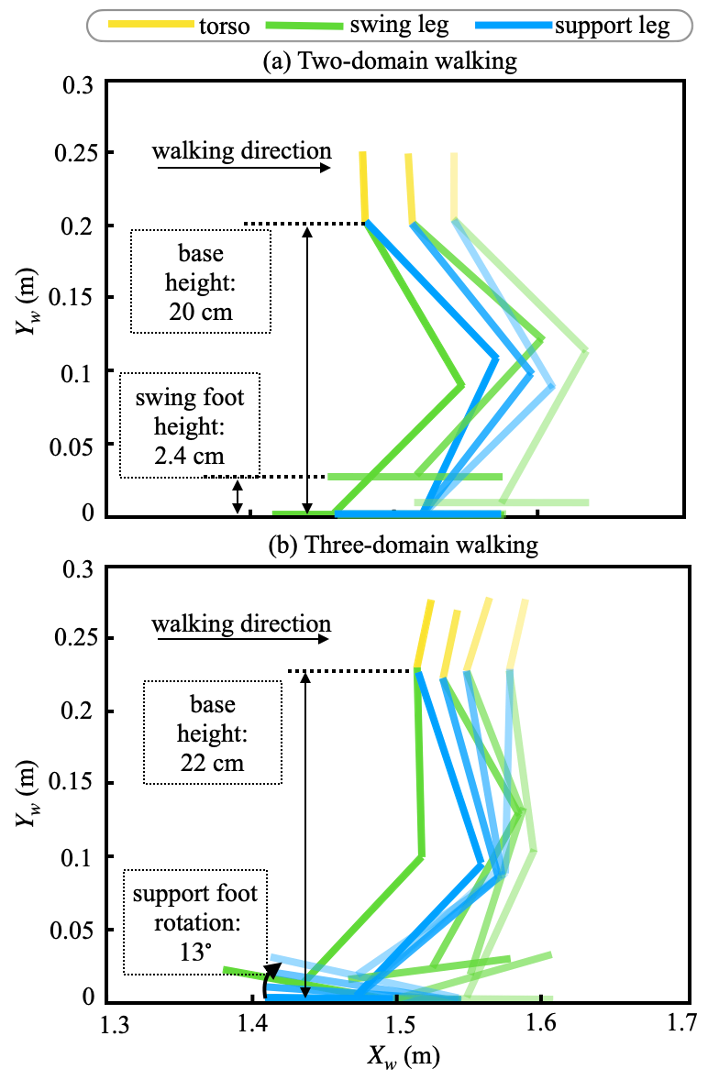}
    \vspace{-0.2 in}
    \caption{Desired walking patterns for (a) two-domain walking (Cases A and B) and (b) three-domain walking (Case C) in the sagittal plane.
    The labels $X_w$ and $Y_w$ represent the $x$- and $y$-axes of the world frame, respectively.}
    \label{fig:walking illustration}
\end{figure}

The desired functions
$\boldsymbol{\phi}^F$, $\boldsymbol{\phi}^U$, and $\boldsymbol{\phi}^O$ are designed as B\'ezier curves (Section 3.2).
To respect the impact dynamics as prescribed by condition (B2), their parameters could be designed using the methods introduced in~\cite{westervelt2007feedback}.
The desired walking patterns corresponding to the desired functions $\boldsymbol{\phi}^F$, $\boldsymbol{\phi}^U$, and $\boldsymbol{\phi}^O$ used in this study are illustrated in Fig.~\ref{fig:walking illustration}.
In three-domain walking (top plot in Fig.~\ref{fig:walking illustration}), the FA, UA, and OA phases take up approximately $33\%$, $8\%$, and $59\%$ of one walking step, respectively, while the FA and UA phases of the two-domain walking gait (lower plot in Fig.~\ref{fig:walking illustration}) last $81\%$ and $19\%$ of a step, respectively.
For both walking patterns, the step length and maximum swing foot height are $7.1~\text{cm}$ and $2.4~\text{cm}$, respectively.

\vspace{-0.2 in}
\subsubsection{Simulation cases}
To validate the proposed controller under different desired global-position trajectories, walking patterns, and initial errors, we simulate the following three cases:
\begin{itemize}
    \item[(Case A):] Combination of desired trajectory (GP1) and
    two-domain walking pattern (Fig.~\ref{fig:walking illustration}, top).
    \item[(Case B):] Combination of desired trajectory (GP2) and two-domain walking pattern (Fig.~\ref{fig:walking illustration}, top).
    \item[(Case C):] 
    Combination of desired trajectory (GP3) and three-domain walking pattern (Fig.~\ref{fig:walking illustration}, bottom).
\end{itemize}

Table~\ref{Table: initial error} summarizes the initial tracking error norms for all cases.
Note that the initial swing-foot position tracking error is roughly 30-40$\%$ of the nominal step length.

\vspace{-0.2 in}
\subsubsection{Controller setting}

For the IO-PD and IO-QP controllers, the PD controller gains are set as $\mathbf{K}_{p,i}=225 \cdot \mathbf{I}$ and $\mathbf{K}_{d,i}=50 \cdot \mathbf{I}$ to ensure the matrix $\bA_i$ ($i \in \{F, U, O \}$) is Hurwitz.
For the IO-QP controller, the slack penalty weight $p$ (Eq.~\eqref{equ: IO-QP}) is set as $p=10^{7}$.
On a desktop with an i7 CPU and 32GB RAM running MATLAB, it takes approximately 1 ms to solve the QP problem in Eq.~\eqref{equ: IO-QP}.

To verify the stability of the multi-domain walking system, we construct the three Lyapunov-like functions $V_f$, $V_u$, and $V_O$ as introduced in Section~\ref{Section-Stability}.
In all domains, the matrix $\mathbf{P}_i$ (where $i \in \{F, U, O\}$) is obtained by solving the Lyapunov equation using the gain matrices $\mathbf{K}_{p,i}$ and $\mathbf{K}_{d,i}$ and the matrix $\mathbf{Q}_i$.
Here without loss of generality, we choose $\mathbf{Q}_i$ as an identity matrix.
For the UA phase, the value of $\beta$ in the definition of $V_U$ in Eq.~\eqref{Vu-def} is set as $0.001$.

\subsection{Simulation Results}
This subsection presents the tracking results of our proposed IO-PD and IO-QP controller for Cases A through C.

\vspace{-0.2 in}
\subsubsection{Global-position tracking performance}

\begin{figure*}[t]
    \centering    \includegraphics[width=1\linewidth]{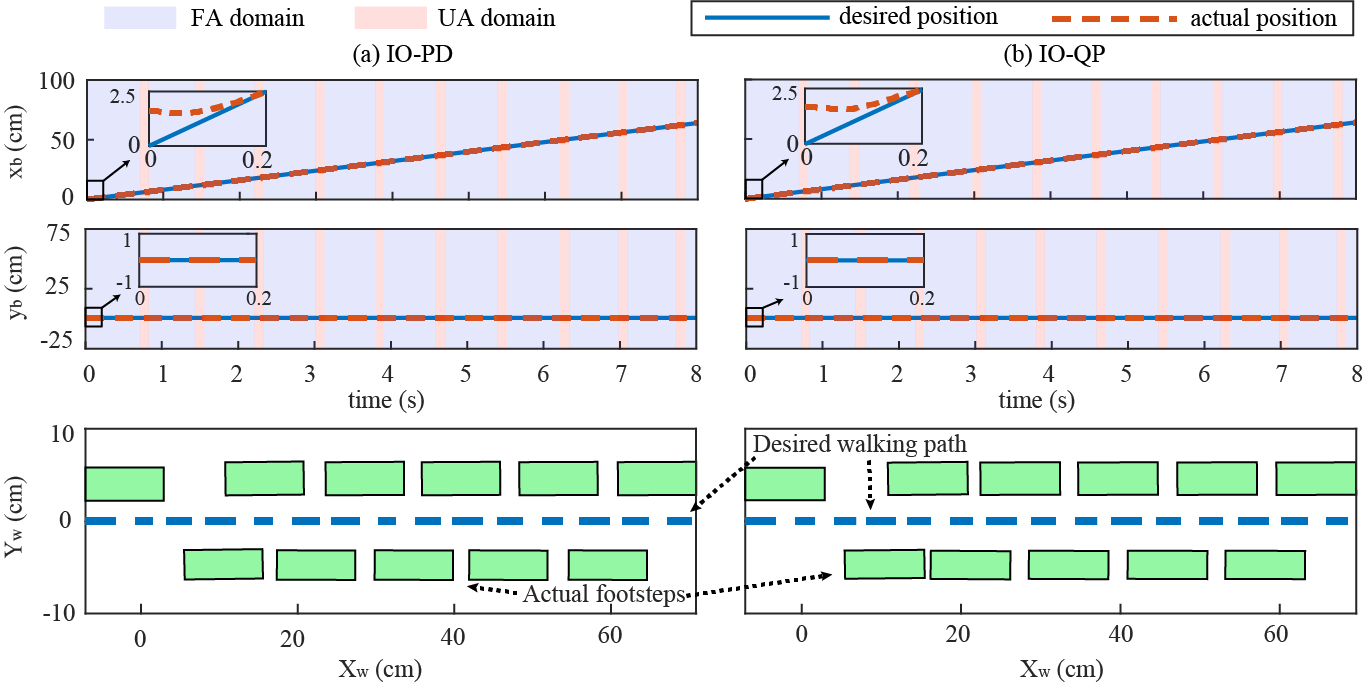}
    \vspace{-0.2in}
    \caption{Satisfactory global-position tracking performance under Case A. 
    The top row shows the global-position tracking results, and the bottom row displays the straight-line desired walking path and the actual footstep locations.
    The initial errors are listed in the Table~\ref{Table: initial error}.
    }
    \label{fig:caseA}
\end{figure*}

\begin{figure*}[t]
    \centering
    \includegraphics[width=1\linewidth]{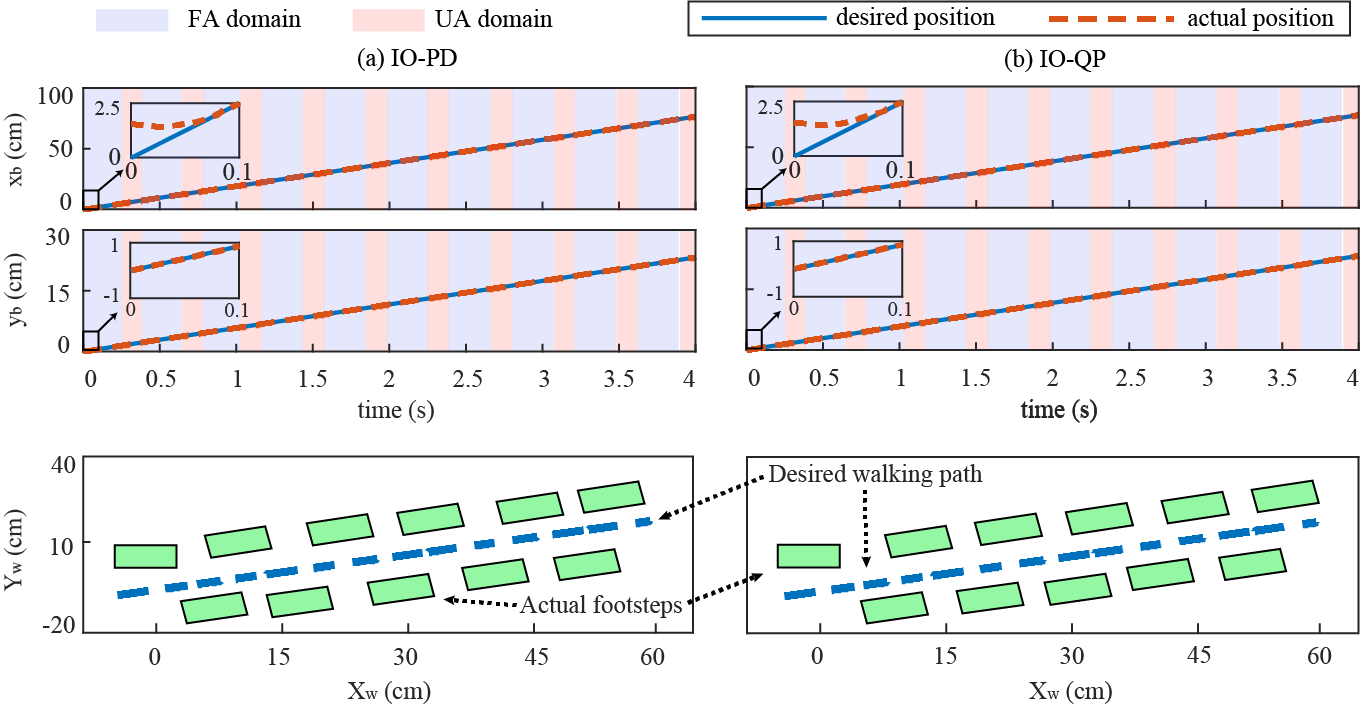}
    \vspace{-0.2 in}
    \caption{Satisfactory global-position tracking performance under Case B. The top row shows the global-position tracking results, and the bottom row displays the desired straight-line walking path and the actual footstep locations.
    The initial errors are listed in the Table~\ref{Table: initial error}.
    }
    \label{fig:caseB}
\end{figure*}

Figures~\ref{fig:caseA} and~\ref{fig:caseB} show the tracking performance of the proposed IO-PD and IO-QP controllers under Cases A and B, respectively.
As explained earlier, Cases A and B share the same desired walking pattern of two-domain walking, but they have different desired global-position trajectories and initial errors.
For both cases, the IO-PD and IO-QP controllers satisfactorily drive the robot's actual horizontal global position $(x_b,y_b)$ to the desired trajectories $(x_d(t),y_d(t))$, as shown in the top four plots in each figure.
Also, from the footstep locations displayed at the bottom of each figure, the robot is able to walk along the desired walking path over the ground.
In particular, the footstep trajectories in Fig.~\ref{fig:caseB} demonstrate that even with a notable initial error (approx. $17^{\degree}$) of the robot's heading direction, the robot is able to quickly converge to the desired walking path.

\begin{figure*}
    \centering
    \includegraphics[width=1\linewidth]{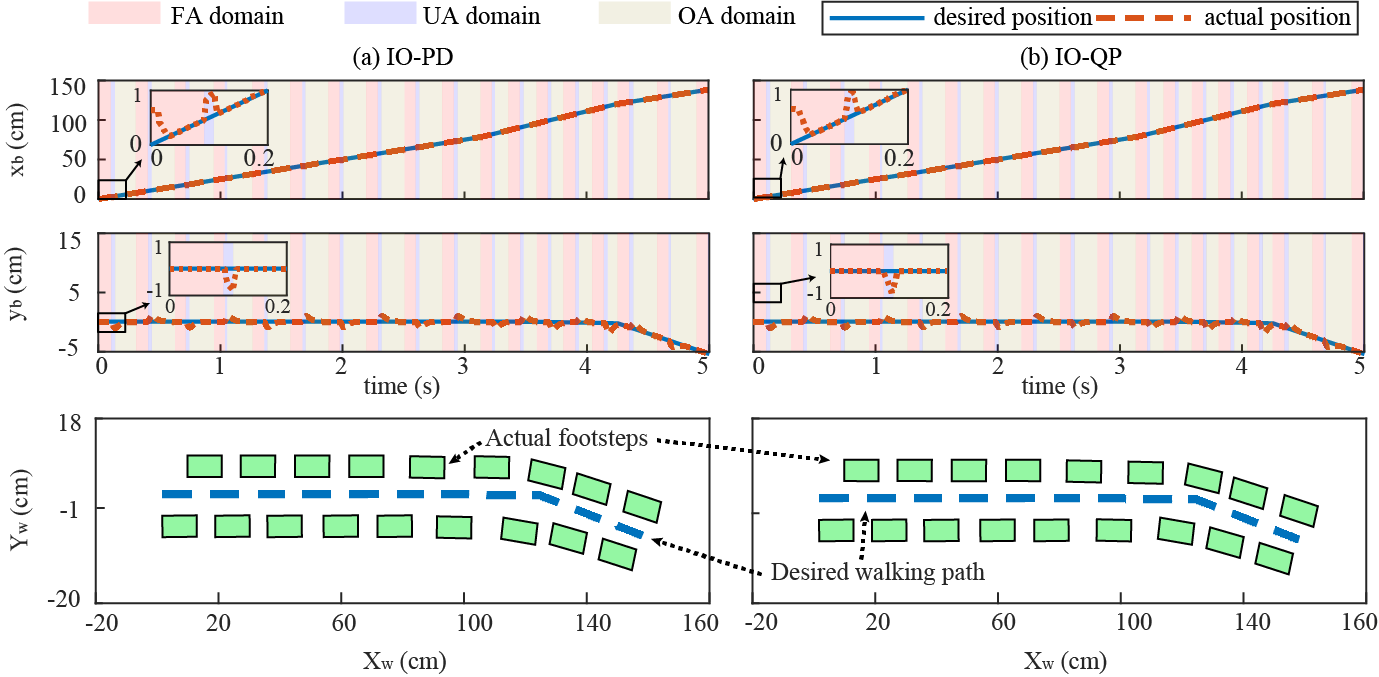}
    \vspace{-0.2 in}
    \caption{Satisfactory global-position tracking performance under Case C.
    The top row shows the global-position tracking results, and the bottom row displays the desired walking path and the actual footstep locations.
    The desired walking path consists of two straight lines connected by an arc.
    The initial errors are listed in the Table~\ref{Table: initial error}.}
    %
    \label{fig:caseC}
\end{figure*}

Figure~\ref{fig:caseC} displays the global-position tracking results of three-domain walking for Case C.
The top two plots, i.e., the time profiles of the forward and lateral base position ($x_b$ and $y_b$), show that the actual horizontal global position diverges from the reference within the UA phase during which the global position is not directly controlled.
Despite the error divergence within the UA phase, the actual global position still converges to close to zero over the entire walking process thanks to convergence within the FA and OA domains, confirming the validity of Theorem 1.

\vspace{-0.2 in}
\subsubsection{Convergence of Lyapunov-like functions} 

\begin{figure}[t]
    \centering
    \includegraphics[width=1\linewidth]{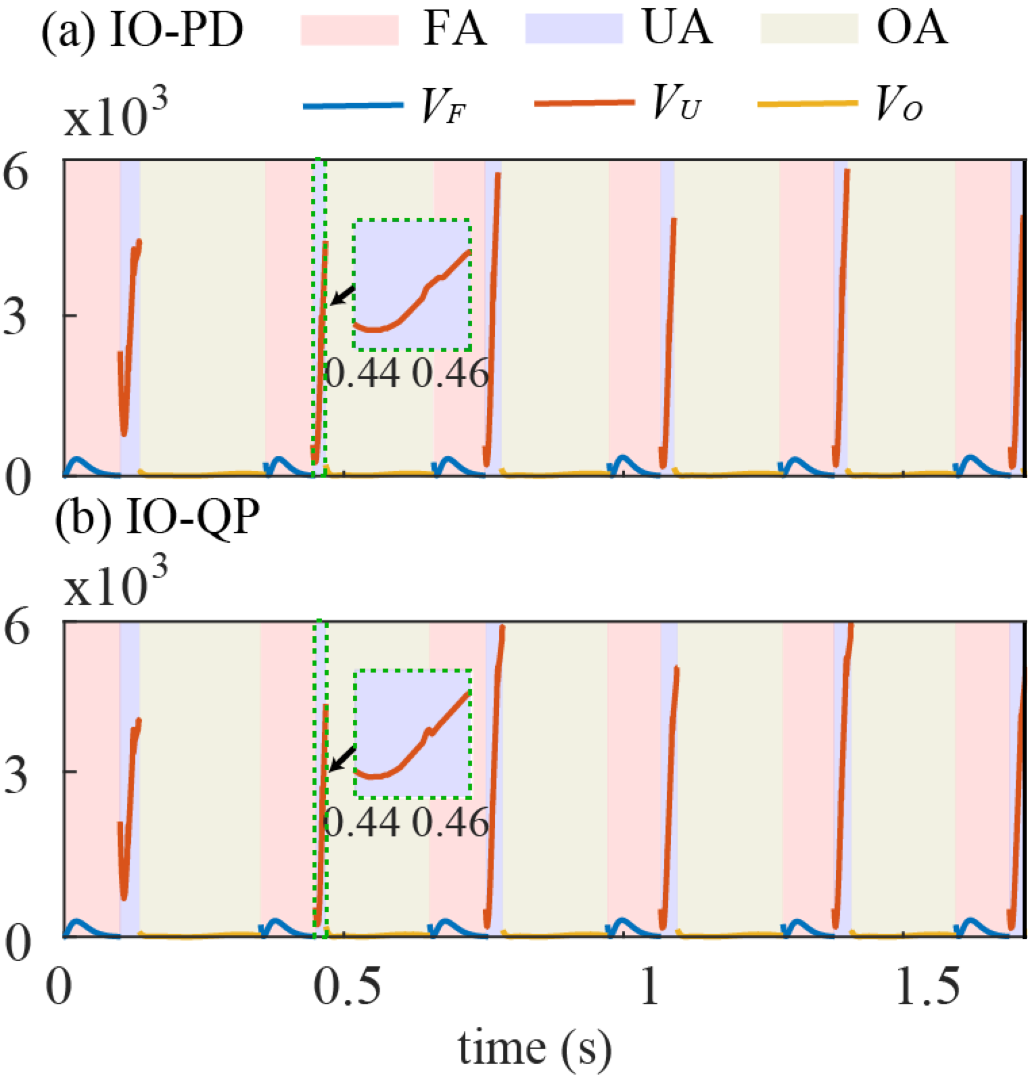}
    \vspace{-0.2in}
    \caption{Time evolutions of multiple Lyapunov-like functions under Case C. 
    The closed-loop stability is confirmed by the behaviors of the multiple Lyapunov functions, which complies with conditions (C1)-(C3) stated in the proof of Theorem 1 for both (a) IO-PD and (b) IO-QP control laws.}
    \label{fig:Lyapunov}
\end{figure}

The multiple Lyapunov-like functions for case C, implemented with IO-PD and IO-QP control laws, is illustrated in Figure~\ref{fig:Lyapunov}.
Both control laws ensure the continuous-phase convergence of $V_F$ and $V_O$ satisfies condition (B1).
Although $V_U$ diverges during the UA phase, it remains bounded, thereby satisfying condition (B3).
Moreover, we know the desired trajectories parameterized as B\'ezier curves are planned to satisfy (B2).
Therefore, the multiple Lyapunov-like functions behave as predicted by conditions (C1)-(C3) in the proof of Theorem 1, indicating closed-loop stability.

\vspace{-0.2 in}
\subsubsection{Satisfaction of torque limits} 
\begin{figure*}[t]
    \centering
\includegraphics[width=1\linewidth]{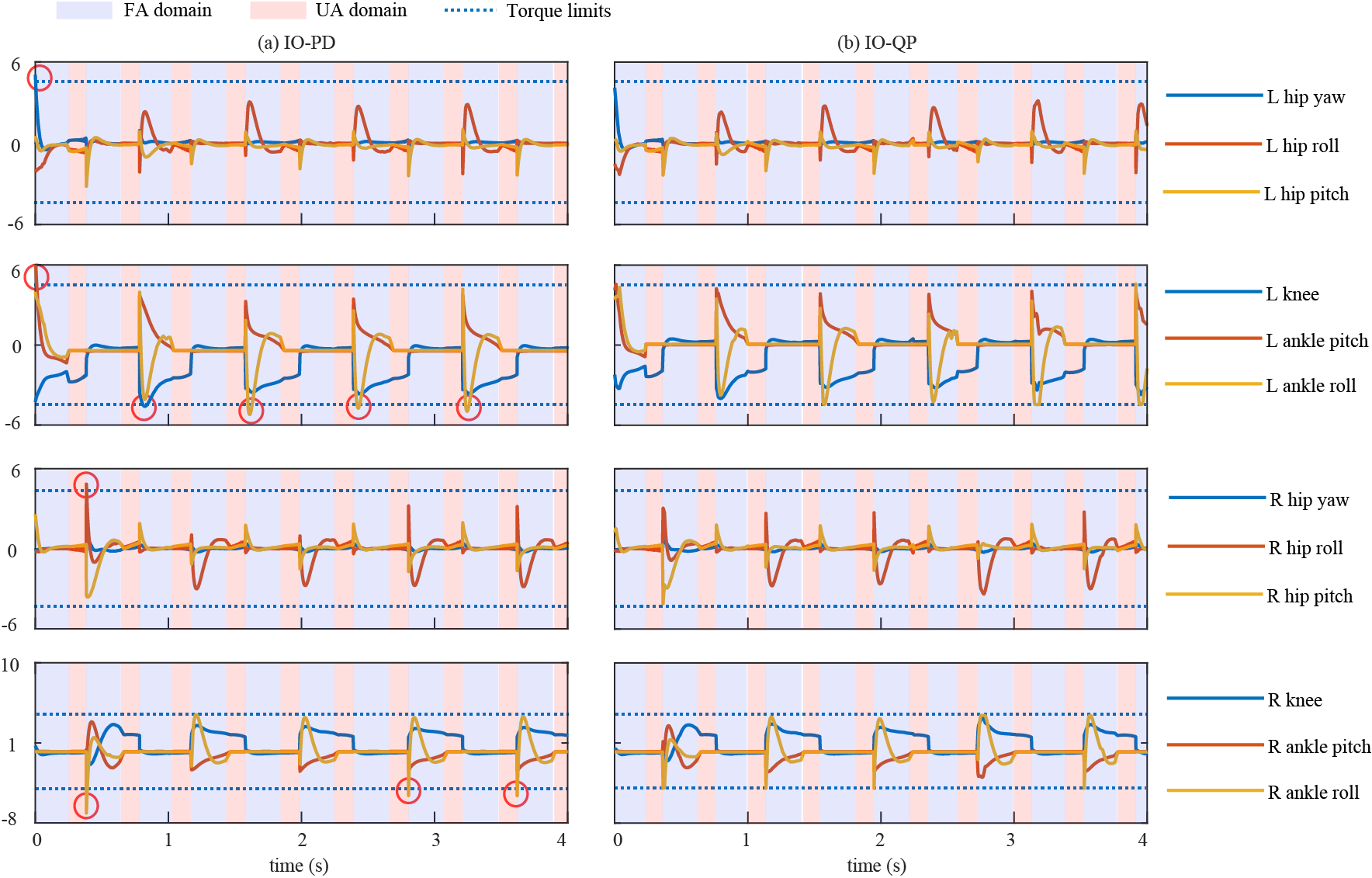}
\vspace{-0.2in}
    \caption{Torque profiles of each leg motor under the proposed (a) IO-PD and (b) IO-QP controllers for Case B.
    ``L'' and ``R'' stand for left and right, respectively.
    The red circles highlight the occurrence of torque limit violations.
    The jumps are more significant under the IO-PD controller than the IO-QP controller because the latter explicitly meets the torque limits. 
    The blue dotted line represents the torque limits. It is evident that the torque profile of the IO-QP controller adheres to the torque limits, whereas the torque profile of the IO-PD controller may exceed the torque limits.
}
    \label{fig:torque}
\end{figure*}

Figure~\ref{fig:torque} illustrates the joint torque profiles of each leg motor under the IO-PD and IO-QP control methods for Case B.
The torque limits $u_{max}$ and $u_{min}$ are set as $4.1$ N and $-4.1$ N, respectively.
It is observed that the torque experiences sudden spikes due to the foot-landing impact at the switching from the UA to the FA phases.
Due to the notable initial tracking errors, there are also multiple spikes in the joint torques at the beginning of the entire walking process.
These spikes tend to be more significant with the IO-PD controller than with the IO-QP controller. 
In fact, all of the torque peaks under IO-QP are within the torque limits whereas some of those peaks under IO-PD exceed the limits, which is primarily due to the fact that the IO-QP controller explicitly enforces the torque limits but IO-PD does not.
This comparison highlights the advantage of using IO-QP over IO-PD in ensuring satisfaction of actuation constraints.

\section{Discussion}
\label{Sec: Discussion}

This study has introduced a nonlinear GPT control approach for 3-D multi-domain bipedal robotic walking based on hybrid full-order dynamics modeling and multiple Lyapunov stability analysis.
Similar to the HZD-based approaches~\cite{hereid2014dynamic,reher2020algorithmic,hamed2019dynamically} for multi-domain walking, our controller only acts within continuous phases, leaving the discrete impact dynamics uncontrolled.
Another key similarity lies in that we also build the controller based on the hybrid, nonlinear, full-order dynamics model of multi-domain walking that faithfully captures the true robot dynamics
and we exploit the input-output linearization technique to exactly linearize the complex continuous-phase robot dynamics.

Despite these similarities, our control law focuses on accurately tracking the desired global-position trajectories with the precise timing, whereas the HZD-based approach may not be directly extended to achieve such global-position tracking performance.
This is essentially caused by the different stability types that the two approaches impose. 
The stability conditions proposed in this study enforce the stability of the desired global-position trajectory, which is a time function encoded by the global time.
In contrast, the stability conditions underlying the HZD framework ensure the stability of the desired periodic orbit, which is a curve in the state space on which infinitely many global-position trajectories reside.

Our previous GPT controller design~\cite{gu2016bipedal} for the multi-domain walking of a 2-D robot is only capable of tracking straight-line paths.
By explicitly modeling the robot dynamics associated with 3-D walking and considering the robot's 3-D movement in the design of the desired trajectories, the proposed approach is capable of ensuring satisfactory global-position tracking performance for 3-D walking.

One limitation of the proposed approach is that it may be non-feasible to meet the proposed stability conditions in practice if the duration of the underactuation phase, $\delta_{\tau_U}$, is overly large.
From Eq.~\eqref{Eq: FContDelta} in the proof of Proposition 3, we know that as $\delta_{\tau_U}$ increases, $\alpha_2$ will also increase, leading to a larger value of $\Bar{N}$.
If $\Bar{N}$ is overly large, Eq.~\eqref{eq: lyapunov boundedness} will no longer hold, and the stability conditions will be invalid.
To resolve this potential issue, the nominal duration of the UA domain cannot be set overly long. 
Indeed, the percentage of the UA phase within a complete gait cycle is respectively $19\%$ and $8 \%$ of the simulated three-domain and two-domain walking,
which is comparable to that of human walking (i.e., $18\%$~\cite{reher2020algorithmic}).

Another limitation of our control laws lies in that the robot dynamics model needs to be sufficiently accurate for the controller to be effective, due to the utilization of the input-output linearization technique.
Yet, model parametric errors, external disturbances, and hardware imperfections (e.g., sensor noise) are prevalent in real-world robot operations~\cite{yeatman2019decentralized}. To enhance the robustness of the proposed controller for real-world applications, we can incorporate robust control~\cite{hu2011experimental,liao2017high,yuan2019fast,gu2021adaptive} into the GPT control law to address uncertainties. 
Furthermore, we can exploit online footstep planning~\cite{gao2022time,iqbalasymptotic,dai2022bipedal,xiong20223,nguyen2020dynamic,gong2022zero} to adjust the robot's desired behaviors in real-time to better reject modeling errors and external disturbances. 

\section{CONCLUSION}
\label{Sec: Conclusion}

This paper has introduced a continuous tracking control law that achieves provably accurate global-position tracking for the hybrid model of multi-domain bipedal robotic walking involving different actuation types.
The proposed control law was derived based on input-output linearization and proportional-derivative control, ensuring the exponential stability of the output function dynamics within each continuous phase of the hybrid walking process. 
Sufficient stability conditions were established via the construction of multiple Lyapunov functions and could be used to guide the gain tuning of the proposed control law for ensuring the provable stability for the overall hybrid system.
Both a three-domain and a two-domain walking gait were investigated to illustrate the effectiveness of the proposed approach, and the input-output linearizing controller was cast into a quadratic program (QP) to handle the actuator torque saturation.
Simulation results on a three-dimensional bipedal humanoid robot confirmed the validity of the proposed control law under a variety of walking paths, desired global-position trajectories, desired walking patterns, and initial errors. 
Finally, the performance of the input-output linearizing control law with and without the QP formulation was compared to highlight the effectiveness of the former in mitigating torque saturation while ensuring the closed-loop stability and trajectory tracking accuracy.

\begin{acknowledgment}
The authors would like to thank Sushant Veer and Ayonga Hereid for their constructive comments on the theory and simulations of this work.
\end{acknowledgment}

%
% ...

\bibliographystyle{asmems4}

\bibliography{asme2e}

\begin{thebibliography}{10}

\bibitem{zhao2017multi}
Zhao, H., Hereid, A., Ma, W.-l., and Ames, A.~D., 2017,
\newblock ``Multi-contact bipedal robotic locomotion,''
\newblock {\em Robotica, {\bf 35}}(5), pp.~1072--1106.

\bibitem{zhao2014human}
Zhao, H.-H., Ma, W.-L., Ames, A.~D., and Zeagler, M.~B., 2014,
\newblock ``Human-inspired multi-contact locomotion with amber2,''
\newblock In Proc. of ACM/IEEE International Conference on Cyber-Physical
  Systems, pp.~199--210.

\bibitem{robotis}
{R}{O}{B}{O}{T}{I}{S} {C}o., {L}td.
\newblock https://www.robotis.us/
\newblock Accessed: 2023-01-20.

\bibitem{Ramezani2014}
Ramezani, A., Hurst, J.~W., Hamed, K.~A., and Grizzle, J.~W., 2014,
\newblock ``Performance analysis and feedback control of {ATRIAS}, a
  three-dimensional bipedal robot,''
\newblock {\em ASME Journal of Dynamic Systems, Measurement, and Control, {\bf
  136}}(2), p.~021012.

\bibitem{schwind1998spring}
Schwind, W.~J., 1998,
\newblock {\em Spring loaded inverted pendulum running: A plant model}
\newblock University of Michigan.

\bibitem{hereid2014dynamic}
Hereid, A., Kolathaya, S., Jones, M.~S., Van~Why, J., Hurst, J.~W., and Ames,
  A.~D., 2014,
\newblock ``Dynamic multi-domain bipedal walking with atrias through {SLIP}
  based human-inspired control,''
\newblock In Proc. of International Conference on Hybrid Systems: Computation
  and Control, pp.~263--272.

\bibitem{grimes2012design}
Grimes, J.~A., and Hurst, J.~W., 2012,
\newblock ``The design of atrias 1.0 a unique monopod, hopping robot,''
\newblock In {\em Adaptive Mobile Robotics}. World Scientific, pp.~548--554.

\bibitem{westervelt2007feedback}
Westervelt, E.~R., Chevallereau, C., Choi, J.~H., Morris, B., and Grizzle,
  J.~W., 2007,
\newblock {\em Feedback control of dynamic bipedal robot locomotion}
\newblock CRC press.

\bibitem{reher2016realizing}
Reher, J., Cousineau, E.~A., Hereid, A., Hubicki, C.~M., and Ames, A.~D., 2016,
\newblock ``Realizing dynamic and efficient bipedal locomotion on the humanoid
  robot {DURUS},''
\newblock In Proc. of IEEE International Conference on Robotics and Automation,
  pp.~1794--1801.

\bibitem{hamed2019dynamically}
Hamed, K.~A., Ma, W.-L., and Ames, A.~D., 2019,
\newblock ``Dynamically stable {3D} quadrupedal walking with multi-domain
  hybrid system models and virtual constraint controllers,''
\newblock In Proc. of American Control Conference, pp.~4588--4595.

\bibitem{akbari2019dynamic}
Hamed, K., Safaee, B., and Gregg, R.~D., 2019,
\newblock ``Dynamic output controllers for exponential stabilization of
  periodic orbits for multidomain hybrid models of robotic locomotion,''
\newblock {\em ASME Journal of Dynamic Systems, Measurement, and Control, {\bf
  141}}(12).

\bibitem{grizzle2001asymptotically}
Grizzle, J.~W., Abba, G., and Plestan, F., 2001,
\newblock ``Asymptotically stable walking for biped robots: Analysis via
  systems with impulse effects,''
\newblock {\em IEEE Transactions on Automatic Control, {\bf 46}}(1),
  pp.~51--64.

\bibitem{westervelt2003hybrid}
Westervelt, E.~R., Grizzle, J.~W., and Koditschek, D.~E., 2003,
\newblock ``Hybrid zero dynamics of planar biped walkers,''
\newblock {\em IEEE Transactions on Automatic Control, {\bf 48}}(1),
  pp.~42--56.

\bibitem{sreenath2011compliant}
Sreenath, K., Park, H.-W., Poulakakis, I., and Grizzle, J.~W., 2011,
\newblock ``A compliant hybrid zero dynamics controller for stable, efficient
  and fast bipedal walking on {MABEL},''
\newblock {\em The International Journal of Robotics Research, {\bf 30}}(9),
  pp.~1170--1193.

\bibitem{veer2019input}
Veer, S., Poulakakis, I., et~al., 2019,
\newblock ``Input-to-state stability of periodic orbits of systems with impulse
  effects via poincar{\'e} analysis,''
\newblock {\em IEEE Transactions Automatic Control, {\bf 64}}(11),
  pp.~4583--4598.

\bibitem{gong2019feedback}
Gong, Y., Hartley, R., Da, X., Hereid, A., Harib, O., Huang, J.-K., and
  Grizzle, J., 2019,
\newblock ``Feedback control of a cassie bipedal robot: Walking, standing, and
  riding a segway,''
\newblock In Proc. of American Control Conference, pp.~4559--4566.

\bibitem{gu2016bipedal}
Gu, Y., Yao, B., and Lee, C.~G., 2016,
\newblock ``Bipedal gait recharacterization and walking encoding generalization
  for stable dynamic walking,''
\newblock In Proc. of IEEE International Conference on Robotics and Automation,
  pp.~1788--1793.

\bibitem{gu2018straight}
Gu, Y., Yao, B., and Lee, C.~G., 2018,
\newblock ``Straight-line contouring control of fully actuated 3-{D} bipedal
  robotic walking,''
\newblock In Proc. of American Control Conference, pp.~2108--2113.

\bibitem{gu2017time}
Gu, Y., Yao, B., and Lee, C.~G., 2017,
\newblock ``Time-dependent orbital stabilization of underactuated bipedal
  walking,''
\newblock In Proc. of American Control Conference, pp.~4858--4863.

\bibitem{gao2019global}
Gao, Y., and Gu, Y., 2019,
\newblock ``Global-position tracking control of a fully actuated {NAO} bipedal
  walking robot,''
\newblock In Proc. of American Control Conference, pp.~4596--4601.

\bibitem{gu2020adaptive}
Gu, Y., and Yuan, C., 2020,
\newblock ``Adaptive robust trajectory tracking control of fully actuated
  bipedal robotic walking,''
\newblock In Proc. of IEEE/ASME International Conference on Advanced
  Intelligent Mechatronics, pp.~1310--1315.

\bibitem{gu2021adaptive}
Gu, Y., and Yuan, C., 2021,
\newblock ``Adaptive robust tracking control for hybrid models of
  three-dimensional bipedal robotic walking under uncertainties,''
\newblock {\em ASME Journal of Dynamic Systems, Measurement, and Control, {\bf
  143}}(8).

\bibitem{gu2022global}
Gu, Y., Gao, Y., Yao, B., and Lee, C.~G., 2022,
\newblock ``Global-position tracking control for three-dimensional bipedal
  robots via virtual constraint design and multiple {Lyapunov} analysis,''
\newblock {\em ASME Journal of Dynamic Systems, Measurement, and Control, {\bf
  144}}(11), p.~111001.

\bibitem{iqbaloptimization}
Iqbal, A., and Gu, Y., 2021,
\newblock ``Extended capture point and optimization-based control for
  quadrupedal robot walking on dynamic rigid surfaces,''
\newblock {\em IFAC-PapersOnLine, {\bf 54}}(20).

\bibitem{9108552}
Iqbal, A., Gao, Y., and Gu, Y., 2020,
\newblock ``Provably stabilizing controllers for quadrupedal robot locomotion
  on dynamic rigid platforms,''
\newblock {\em IEEE/ASME Transactions on Mechatronics, {\bf 25}}(4),
  pp.~2035--2044.

\bibitem{iqbal2023real}
Iqbal, A., Veer, S., and Gu, Y., 2023,
\newblock ``Real-time walking pattern generation of quadrupedal dynamic-surface
  locomotion based on a linear time-varying pendulum model,''
\newblock {\em arXiv preprint arXiv:2301.03097}.

\bibitem{gao_multi_2019global}
Gao, Y., and Gu, Y., 2019,
\newblock ``Global-position tracking control of multi-domain planar bipedal
  robotic walking,''
\newblock In Prof. of ASME Dynamic Systems and Control Conference, Vol.~59148,
  p.~V001T03A009.

\bibitem{bhounsule2017discrete}
Bhounsule, P.~A., and Zamani, A., 2017,
\newblock ``A discrete control lyapunov function for exponential orbital
  stabilization of the simplest walker,''
\newblock {\em Journal of Mechanisms and Robotics, {\bf 9}}(5).

\bibitem{khalil1996noninear}
Khalil, H.~K., 1996,
\newblock {\em Noninear systems}
\newblock No.~5. Prentice Hall.

\bibitem{chan2018optimization}
Chan, W.~K., Gu, Y., and Yao, B., 2018,
\newblock ``Optimization of output functions with nonholonomic virtual
  constraints in underactuated bipedal walking control,''
\newblock In Proc. of Annual American Control Conference, pp.~6743--6748.

\bibitem{branicky1998multiple}
Branicky, M.~S., 1998,
\newblock ``Multiple lyapunov functions and other analysis tools for switched
  and hybrid systems,''
\newblock {\em IEEE Transactions on Automatic Control, {\bf 43}}(4),
  pp.~475--482.

\bibitem{khalil1996nonlinear}
Khalil, H.~K., 1996,
\newblock {\em Nonlinear control}
\newblock Prentice Hall.

\bibitem{gu2017time_thesis}
Gu, Y., 2017,
\newblock ``Time-dependent nonlinear control of bipedal robotic walking,''
\newblock PhD thesis, Purdue University.

\bibitem{rijnen2019hybrid}
Rijnen, M., Biemond, J.~B., Van De~Wouw, N., Saccon, A., and Nijmeijer, H.,
  2019,
\newblock ``Hybrid systems with state-triggered jumps: Sensitivity-based
  stability analysis with application to trajectory tracking,''
\newblock {\em IEEE Transactions on Automatic Control, {\bf 65}}(11),
  pp.~4568--4583.

\bibitem{rijnen2016hybrid}
Rijnen, M., van Rijn, A., Dallali, H., Saccon, A., and Nijmeijer, H., 2016,
\newblock ``Hybrid trajectory tracking for a hopping robotic leg,''
\newblock {\em IFAC-PapersOnLine, {\bf 49}}(14), pp.~107--112.

\bibitem{gong2020angular}
Gong, Y., and Grizzle, J., 2020,
\newblock ``Angular momentum about the contact point for control of bipedal
  locomotion: Validation in a {LIP}-based controller,''
\newblock {\em arXiv preprint arXiv:2008.10763}.

\bibitem{reher2020algorithmic}
Reher, J.~P., Hereid, A., Kolathaya, S., Hubicki, C.~M., and Ames, A.~D., 2020,
\newblock ``Algorithmic foundations of realizing multi-contact locomotion on
  the humanoid robot {DURUS},''
\newblock In Algorithmic Foundations of Robotics XII: Proceedings of the
  Twelfth Workshop on the Algorithmic Foundations of Robotics, Springer,
  pp.~400--415.

\bibitem{yeatman2019decentralized}
Yeatman, M., Lv, G., and Gregg, R.~D., 2019,
\newblock ``Decentralized passivity-based control with a generalized energy
  storage function for robust biped locomotion,''
\newblock {\em ASME Journal of Dynamic Systems, Measurement, and Control, {\bf
  141}}(10).

\bibitem{hu2011experimental}
Hu, C., Yao, B., Wang, Q., Chen, Z., and Li, C., 2011,
\newblock ``Experimental investigation on high-performance coordinated motion
  control of high-speed biaxial systems for contouring tasks,''
\newblock {\em International Journal of Machine Tools and Manufacture, {\bf
  51}}(9), pp.~677--686.

\bibitem{liao2017high}
Liao, J., Chen, Z., and Yao, B., 2017,
\newblock ``High-performance adaptive robust control with balanced torque
  allocation for the over-actuated cutter-head driving system in tunnel boring
  machine,''
\newblock {\em Mechatronics, {\bf 46}}, pp.~168--176.

\bibitem{yuan2019fast}
Yuan, M., Chen, Z., Yao, B., and Liu, X., 2019,
\newblock ``Fast and accurate motion tracking of a linear motor system under
  kinematic and dynamic constraints: an integrated planning and control
  approach,''
\newblock {\em IEEE Transactions on Control System Technology}.

\bibitem{gao2022time}
Gao, Y., Gong, Y., Paredes, V., Hereid, A., and Gu, Y., 2022,
\newblock ``Time-varying {ALIP} model and robust foot-placement control for
  underactuated bipedal robot walking on a swaying rigid surface,''
\newblock {\em arXiv preprint arXiv:2210.13371}.

\bibitem{iqbalasymptotic}
Iqbal, A., Veer, S., and Gu, Y., 2023,
\newblock ``Asymptotic stabilization of aperiodic trajectories of a
  hybrid-linear inverted pendulum walking on a dynamic rigid surface,''
\newblock In Proc. of American Control Conference, to appear.

\bibitem{dai2022bipedal}
Dai, M., Xiong, X., and Ames, A., 2022,
\newblock ``Bipedal walking on constrained footholds: Momentum regulation via
  vertical com control,''
\newblock In Proc. of IEEE International Conference on Robotics and Automation,
  pp.~10435--10441.

\bibitem{xiong20223}
Xiong, X., and Ames, A., 2022,
\newblock ``3-{D} underactuated bipedal walking via {H-LIP} based gait
  synthesis and stepping stabilization,''
\newblock {\em IEEE Transactions on Robotics, {\bf 38}}(4), pp.~2405--2425.

\bibitem{nguyen2020dynamic}
Nguyen, Q., Da, X., Grizzle, J., and Sreenath, K., 2020,
\newblock ``Dynamic walking on stepping stones with gait library and control
  barrier functions,''
\newblock In Algorithmic Foundations of Robotics XII: Proceedings of the
  Twelfth Workshop on the Algorithmic Foundations of Robotics, pp.~384--399.

\bibitem{gong2022zero}
Gong, Y., and Grizzle, J.~W., 2022,
\newblock ``Zero dynamics, pendulum models, and angular momentum in feedback
  control of bipedal locomotion,''
\newblock {\em ASME Journal of Dynamic Systems, Measurement, and Control, {\bf
  144}}(12), p.~121006.

\end{thebibliography}

\appendix       
\section{Appendix: Proofs of Propositions and Theorem 1}

\label{proof}

\subsection{Proof of Proposition~\ref{prop: 2}}
\label{Appendix: Prop2}
Integrating both sides of the UA closed-loop dynamics in Eq.~\eqref{eq: xu dynamics} over time $t$ yields
\vspace{-0.2 in}
\begin{equation}
\vspace{-0.2 in}
    \mathbf{x}_{\eta} |^-_{3k-1}
    =
    \int^{T_{3k-1}}_{T_{3k-2}} \mathbf{f}_{\eta} (s,\mathbf{x}_{\eta}(s),\mathbf{x}_{\xi}(s)) d s 
    +
    \mathbf{x}_{\eta} |^+_{3k-2} .
\end{equation}
Then,  
\vspace{-0.2 in}
\begin{equation}
\begin{aligned}
     \Big \| \mathbf{x}_{\eta} |^-_{3k-1} \Big \|
    &\leq 
    \Big \| \int^{T_{3k-1}}_{T_{3k-2}} \mathbf{f}_{\eta} (s,\mathbf{x}_{\eta}(s),\mathbf{x}_{\xi}(s)) d s \Big \| 
    +
    \Big \| \mathbf{x}_{\eta} |^+_{3k-2} \Big \| 
    \\
    &\leq 
     \int^{T_{3k-1}}_{T_{3k-2}} \Big \| \mathbf{f}_{\eta} (s,\mathbf{x}_{\eta}(s),\mathbf{x}_{\xi}(s)) \Big \|  d s 
    +
    \Big \| \mathbf{x}_{\eta} |^+_{3k-2} \Big \| .
\end{aligned}
\end{equation}

Since the expression of $\mathbf{f}_{\eta}(\cdot)$ is obtained using
the continuous-phase dynamics of the generalized coordinates in Eq.~\eqref{complete dynamics}
and the expression of the output function $\by_U$ in Eqs.~\eqref{desired walking pattern Over} and \eqref{output function},
we know $\mathbf{f}_{\eta} (t,\mathbf{x}_{\eta},\mathbf{x}_{\xi})$ is continuously differentiable in $t$, $\mathbf{x}_{\eta}$, and $\mathbf{x}_{\xi}$.
Also, we can prove that
there exists a finite, real, positive number $r_{\eta}$ such that
$ \| \frac{\partial \mathbf{f}_{\eta}}{\partial t} \|$,
$ \| \frac{\partial \mathbf{f}_{\eta}}{\partial \mathbf{x}_{\xi}} \|$,
and $\| \frac{\partial \mathbf{f}_{\eta}}{\partial \mathbf{x}_{\eta}} \|$ are bounded on $(T_{3k-2},T_{3k-1}) \times B_{r_\eta} (\mathbf{0})$.
Then, $\mathbf{f}_{\eta} (t,\mathbf{x}_{\eta},\mathbf{x}_{\xi})$ is Lipschitz continuous on $(T_{3k-2},T_{3k-1}) \times B_{r_\eta} (\mathbf{0})$~\cite{khalil1996noninear}, and we can prove that there exists a a real, positive number $k_f$
such that 
\vspace{-0.2 in}
\begin{equation}
\vspace{-0.2 in}
    \Big \| \mathbf{f}_{\eta} (t,\mathbf{x}_{\eta}(t),\mathbf{x}_{\xi}(t)) \Big \| 
    \leq
    k_f
\end{equation}
holds for any $ t \times (\mathbf{x}_{\eta},\mathbf{x}_{\xi} ) \in (T_{3k-2},T_{3k-1}) \times B_{r_\eta} (\mathbf{0})$.

Combining the two inequalities above, we have
\vspace{-0.2 in}
\begin{equation}
\vspace{-0.2 in}
    \Big \|  \mathbf{x}_{\eta} |^-_{3k-1} \Big \| 
    \leq 
    k_f (T_{3k-1}-T_{3k-2})
    + \Big \| \mathbf{x}_{\eta} |^+_{3k-2} \Big \| .
    \label{eq-eta}
\end{equation}
The duration $(T_{3k-1}-T_{3k-2})$ of the UA phase can be estimated as:
\vspace{-0.2 in}
\begin{equation}
\vspace{-0.2 in}
\begin{aligned}
    | T_{3k-1}-T_{3k-2} | &= | T_{3k-1}-\tau_{3k-1}+\tau_{3k-1}-T_{3k-2} | \\
    &\leq 
    | T_{3k-1}-\tau_{3k-1} | +\delta_{\tau_{U}},
\end{aligned}
\end{equation}
where $\delta_{\tau_{U}}:=\tau_{3k-1}-T_{3k-2}$ is the expected duration of the UA phase and $| T_{3k-1}-\tau_{3k-1} |$ is the absolute difference between the actual and planned time instants of the UA$\rightarrow$OA switching.

From our previous work~\cite{gao2019global}, we know there exists small positive numbers $\epsilon_U$ and $r_{U1}$ such that 
\vspace{-0.2 in}
\begin{equation}
\vspace{-0.2 in}
    |T_{3k-1}-\tau_{3k-1} | \leq \epsilon_U \delta_{\tau_{U}}
    \label{eq-tU}
\end{equation}
holds for any $k \in \{1,2,...\}$ and $\mathbf{x}_{U} \in B_{r_{U1}} (\mathbf{0}).$

Thus, using Eqs.~\eqref{eq-eta}-\eqref{eq-tU}, we have
\vspace{-0.2 in}
\begin{equation}
\vspace{-0.2 in}
     \Big \|  \mathbf{x}_{\eta} |^-_{3k-1}  \Big \| 
    \leq 
    k_f ( 1+\epsilon_U ) \delta_{\tau_{U}} 
    +  \Big \|  \mathbf{x}_{\eta} |^+_{3k-2}  \Big \| .
    \label{eq-eta-final}
\end{equation}

Substituting Eqs.~\eqref{eq: Vxi-cont} and \eqref{eq-eta-final} into Eq.~\eqref{Vu-def} gives
\vspace{-0.2 in}
\begin{equation}
\begin{aligned}
    V_U |^-_{3k-1} 
    =& V_{\xi} |^+_{3k-1} + \beta \Big \|\mathbf{x}_{\eta} |^+_{3k-1} \Big \|^2
    \\
    \leq &
    e^{-c_{3\xi} ( T_{3k-1} - T_{3k-2} )} V_{\xi}|^+_{3k-2}
    +
    2 \beta  \Big \|  \mathbf{x}_{\eta} |^+_{3k-2}  \Big \|^2 
    \\
    &+ 2 \beta k_f^2 ( 1+\epsilon_U )^2 \delta_{\tau_{U}}^2 
    \\
    \leq & 2 V_U |^+_{3k-2} 
    + 2 \beta k_f^2 ( 1+\epsilon_U )^2.
\end{aligned}
\label{eq: VU-cont}
\end{equation}
Thus, for any $\mathbf{x}_U \in B_{r_{U2}} (\mathbf{0})$ with $r_{U2} := \min(r_{\eta},r_{U1})$, 
\vspace{-0.2 in}
\begin{equation*}
\vspace{-0.2 in}
    V_U |^-_{3k-1}  \leq w_u(V_U |^+_{3k-2}  )
\end{equation*}
holds,
where
$w_u(V_U |^+_{3k-2}):=2 V_U |^+_{3k-2} 
    + 2 \beta k_f^2 ( 1+\epsilon_U )^2$.
It is clear that $w_u(V_U |^+_{3k-2})$ is a positive-definite function of $V_U |^+_{3k-2}$.
$\hfill
\blacksquare$

\subsection{Proof of Proposition~\ref{prop: 3}}

\label{Appendix: Prop3}

For brevity, we only show the proof for $... \leq  V_F|_{3k}^+  \leq  V_F|_{3k-3}^+ \leq ... \leq V_F|_{3}^+ \leq  V_F|_{0}^+$, based on which the proofs for the other two sets of inequalities in Eq.~\eqref{eq: lyapunov boundedness} can be readily obtained.

To prove that $V_F|_{3k}^+  \leq  V_F|_{3k-3}^+$ for any $k\in \{ 1,2,...\}$, 
we need to analyze the evoluation of the state variables for the $k^{th}$ actual complete gait cycle on $t\in (T_{3k-3},T_{3k})$, which comprises three continuous phases and three switching events.

\noindent \textbf{Analyzing the continuous-phase state evolution:}
We analyze the state evolution during the three continuous phases based on the convergence and boundedness results established in Propositions \ref{prop: 1} and \ref{prop: 2}.

Similar to the boundedness of the UA$\rightarrow$OA switching time discrepancy given in Eq.~\eqref{eq-tU}, there exist small positive numbers $\epsilon_F$, $\epsilon_O$, $r_{tF}$ and $r_{tO}$ such that for any $\bx_F \in \bB_{r_{tF}}(\bzero)$ and $\bx_O \in \bB_{r_{tO}}(\bzero)$,
\vspace{-0.2 in}
\begin{equation}
\vspace{-0.2 in}
    \Big|T_{3k-2}-\tau_{3k-2} \Big| \leq \epsilon_F \delta_{\tau_{F}}
    ~\text{and}~
    \Big| T_{3k}-\tau_{3k} \Big| \leq \epsilon_O \delta_{\tau_{O}}
    \label{eq-tF}
\end{equation}
hold, 
where $\delta_{\tau_{F}}$ and $\delta_{\tau_{O}}$ are the desired periods of the FA and OA phases of the planned walking cycle, with 
$\delta_{\tau_{F}}:=\tau_{3k-2}-T_{3k-3}$ and $\delta_{\tau_{O}}:=\tau_{3k}-T_{3k-1}$ .

Substituting Eq.~\eqref{eq-tF} into Eqs.~\eqref{eq: VF-cont} and \eqref{eq: VO-cont} yields
\vspace{-0.2 in}
\begin{equation}
\vspace{-0.2 in}
    \Big \|  \mathbf{x}_F |^-_{3k-2} \Big \| 
    \leq 
    \sqrt{\tfrac{c_{2F}}{c_{1F}}} 
    e^{-\tfrac{c_{3F}}{2 c_{2F}} (1+\epsilon_F)\delta_{\tau_F} }
    \Big \|  \mathbf{x}_F |^+_{3k-3}  \Big \|
    \label{Eq: ContF-1}
\end{equation}
and
\vspace{-0.2 in}
\begin{equation}
\vspace{-0.2 in}
    \Big \|  \mathbf{x}_O |^-_{3k} \Big \| 
    \leq 
    \sqrt{\tfrac{c_{2O}}{c_{1O}}} 
    e^{-\tfrac{c_{3O}}{2 c_{2O}} (1+\epsilon_O)\delta_{\tau_O} } 
    \Big \|  \mathbf{x}_O  |^+_{3k-1}  \Big \|
    \label{Eq: ContF-2}
\end{equation}
for any 
$\mathbf{x}_i \in B_{\bar{r}_i}(\mathbf{0})$ ($i \in \{F,O \}$),
with the small positive number $\bar{r}_i$ defined as $\bar{r}_i:=\min \{ r_i, r_{ti} \}$.

From the definition of the Lyapunov-like function $V_U$ in Eq.~\eqref{Vu-def}, the continuous-phase boundedness of $V_U$ in Eq.~\eqref{eq: VU-cont}, and the continuous-phase convergence of $V_{\xi}$ in Eq.~\eqref{eq: Vxi-cont}, we obtain the following inequality characterizing the boundedness of the state variable $\bx_U$ within the UA phase:
\vspace{-0.2 in}
\begin{equation}
\vspace{-0.2 in}
\Big\| \bx_U|_{3k-1}^- \Big\|^2
\leq
2
\tfrac{\tilde{c}_{2  \xi}}{\tilde{c}_{1 \xi}}    
\Big\| \bx_U|_{3k-2}^+ \Big\|^2 
+
\tfrac{2 \beta k_f^2}{\tilde{c}_{1 \xi}} 
(1+\epsilon_{U})^2
\label{Eq: ContF-3-2}
\end{equation}
where the real scalar constants $\tilde{c}_{1 \xi}$ and $\tilde{c}_{2 \xi }$ are defined as $\tilde{c}_{1 \xi}:=\min ({c}_{1 \xi },\beta)$ and $\tilde{c}_{2 \xi}:=\max ({c}_{2 \xi},\beta)$.

Since
\vspace{-0.2 in}
\begin{equation*}
\vspace{-0.2 in}
\begin{aligned}
&2 \tfrac{\tilde{c}_{2  \xi}}{\tilde{c}_{1 \xi}}    
\Big \| \bx_U|_{3k-2}^+ \Big \|^2 
+
\tfrac{2 \beta k_f^2}{\tilde{c}_{1 \xi}} 
( 1 + 
\epsilon_{U})^2
\\
\leq &
\left( \sqrt{2 \tfrac{\tilde{c}_{2  \xi}}{\tilde{c}_{1 \xi}} }   
\Big \| \bx_U|_{3k-2}^+ \Big \|
+
\sqrt{\tfrac{2 \beta k_f^2}{\tilde{c}_{1 \xi}} }
( 1 + 
\epsilon_{U}) \right)^2,
\end{aligned}
\end{equation*}
we rewrite Eq.~\eqref{Eq: ContF-3-2} as:
\vspace{-0.2 in}
\begin{equation}
\vspace{-0.2 in}
\begin{aligned}
    \Big \| \bx_U|_{3k-1}^- \Big \|
&\leq
\sqrt{2 \tfrac{\tilde{c}_{2  \xi}}{\tilde{c}_{1 \xi}} }   
\Big \| \bx_U|_{3k-2}^+ \Big \|
+
\sqrt{\tfrac{2 \beta k_f^2}{\tilde{c}_{1 \xi}} }
( 1 + 
\epsilon_{U})
\\
&=: \alpha_1 \Big \| \bx_U|_{3k-2}^+ \Big \| + \alpha_2.
\end{aligned}
    \label{Eq: ContF-3}
\end{equation}

\noindent \textbf{Analyzing the state evolution across a jump:}
Without loss of generality, we first examine the state evolution across the F$\rightarrow$U switching event by relating the norms of the state variable just before and after the impact.

Using the expression of the reset map $\bm \Delta_{F \rightarrow U}$ at the switching instant $t=T_{3k-2}^-$ ($k \in \{ 1,2,...\}$), we obtain the following inequality
\begin{equation}
\small
\begin{aligned}
	\Big \|   \mathbf{x}_U |^+_{3k-2} \Big \| 
	 =& \Big \|   \bm \Delta_{F \rightarrow U} (T_{3k-2},\mathbf{x}_F |^-_{3k-2}  )   \Big \| 
\\
=& 
\Big \|
(\bm \Delta_{F \rightarrow U} 
	(T_{3k-2},\mathbf{x}_F |^-_{3k-2}  ) 
	-  
	\bm \Delta_{F \rightarrow U} 
	(\tau_{3k-2}, \mathbf{x}_F |^-_{3k-2}   ) )
 \\
&+
(  \bm \Delta_{F \rightarrow U} 
	(\tau_{3k-2}, \mathbf{x}_F |^-_{3k-2}   )
	-  \bm \Delta_{F \rightarrow U} 
	(\tau_{3k-2},\mathbf{0}) )
 \\
&+
\bm \Delta_{F \rightarrow U} (\tau_{3k-2},\mathbf{0}) 
\Big \|
 \\
\leq& 
	\Big \|  \bm \Delta_{F \rightarrow U} 
	(T_{3k-2},\mathbf{x}_F |^-_{3k-2}  ) 
	-  
	\bm \Delta_{F \rightarrow U} 
	(\tau_{3k-2}, \mathbf{x}_F |^-_{3k-2}   )   \Big \| 
	\\
	&
	+ 
	\Big \| \bm \Delta_{F \rightarrow U} 
	(\tau_{3k-2}, \mathbf{x}_F |^-_{3k-2}   )
	-  \bm \Delta_{F \rightarrow U} 
	(\tau_{3k-2},\mathbf{0}) \Big \| 
	\\
	&
	+ 
	\Big \|  \bm \Delta_{F \rightarrow U} (\tau_{3k-2},\mathbf{0})   \Big \|.
	\label{Delta_f2u}
\end{aligned}
\end{equation}

Next, we relate the three terms on the right-hand side of the inequality in Eq.~\eqref{Delta_f2u} explicitly with the norm of the state just before the switching (i.e., $\mathbf{x}_F |^-_{3k-2}$).

Recall that the expressions of
$\bm \Delta_{F \rightarrow U}(t,\bx_F)$
solely depends on the expressions of:
(i) the impact dynamics $\boldsymbol{\Delta}_{\dq}(\bq) \dot{\bq}$, which is continuously differentiable on $(\bq, \dot{\bq}) \in \mathcal{T Q}$;
(ii) the output functions $\by_F(t, \bq)$, which is continuously differentiable on $t \in \mathbb{R}^+$ and $\bq \in \mathcal{Q}$ under assumption (A7);
and 
(iii) the time derivative $\dot{\by}_F(t, \bq,\dot{\bq})$, which, also under assumption (A7), is continuously differentiable on $t \in \mathbb{R}^+$ and $(\bq, \dot{\bq}) \in \mathcal{T Q}$.
Thus,
we know
$ \bm \Delta_{F \rightarrow U} $ is continuously differentiable for any $t \in \mathbb{R}^+$ (i.e., including any continuous phases) and state $\mathbf{x}_F \in  \mathbb{R}^{2n}$.

Similarly, under assumption (A7), we can prove that 
there exists a small, real constant $l_{F}$ such that
$ \| \frac{ \partial \bm \Delta_{F \rightarrow U} }{\partial t} \|$
and
$\| \frac{ \partial \bm \Delta_{F \rightarrow U}  }{\partial \mathbf{x}_{F}  } \|$
are bounded for any $t\in \mathbb{R}^+$ (including all continuous FA phases) and $ \bx_F \in B_{l_F} (\mathbf{0})$.
Thus, for any $k \in \{1,2,... \}$,
the function $\bm \Delta_{F \rightarrow U}$ is Lipschitz continuous on
for any $ t \in [T_{3k-2};\tau_{3k-2}]$ and $\bx_F \in B_{l_F} (\mathbf{0})$, where $[T_{3k-2};\tau_{3k-2}]$ equals $[T_{3k-2},\tau_{3k-2}]$ if $T_{3k-2} < \tau_{3k-2}$, and it equals $[\tau_{3k-2},T_{3k-2}]$ if $T_{3k-2} > \tau_{3k-2}$.

Thus, there exist Lipschitz constants $L_{tF}$ and $L_{xF}$ such that:
\begin{equation} 
\small
\begin{aligned}
    &	\Big \|  \bm \Delta_{F \rightarrow U} 
	(T_{3k-2},\mathbf{x}_F |^-_{3k-2}  ) 
	-  
	\bm \Delta_{F \rightarrow U} 
	(\tau_{3k-2}, \mathbf{x}_F |^-_{3k-2}   )   \Big \|  
	\\
	\leq &
	L_{tF} | T_{3k-2} - \tau_{3k-2} |
\end{aligned}
\label{Eq: DeltaF-1}
\end{equation}
and
\vspace{-0.2 in}
\begin{equation} 
\vspace{-0.2 in}
\begin{aligned}
    &\Big \| \bm \Delta_{F \rightarrow U} 
	(\tau_{3k-2}, \mathbf{x}_F |^-_{3k-2}   )
	-  \bm \Delta_{F \rightarrow U} 
	(\tau_{3k-2},\mathbf{0}) \Big \| 
	\\
    \leq & L_{xF} \Big \| \mathbf{x}_F |^-_{3k-2}   \Big \|
\end{aligned}
\label{Eq: DeltaF-2}
\end{equation}
hold on
$[T_{3k-2};\tau_{3k-2}] \times B_{l_F} (\mathbf{0})$ for any $k \in \{1,2,... \}$.

From condition (A2) and Eqs.~\eqref{reset_f2u}, \eqref{eq-tF}, and \eqref{Delta_f2u}-\eqref{Eq: DeltaF-2}, we know that
\vspace{-0.2 in}
\begin{equation}
\vspace{-0.2 in}
    \Big \|   \mathbf{x}_U |^+_{3k-2} \Big \| 
    \leq 
     L_{xF} \Big \| \mathbf{x}_F |^-_{3k-2}   \Big \|
     +L_{tF} \epsilon_F \delta_{\tau F} 
    + \gamma_\Delta
    .
    \label{Eq: DeltaF-3}
\end{equation}

Analogous to the derivation of the inequality in Eq.~\eqref{Eq: DeltaF-3}, we can show that there exist a real, positive number $l_U$ and Lipschitz constants $L_{tU}$ and $L_{xU}$ such that:
\vspace{-0.2 in}
\begin{equation}
\vspace{-0.2 in}
    \Big \|   \mathbf{x}_O |^+_{3k-1} \Big \| 
    \leq 
    L_{xU} \Big \| \mathbf{x}_U |^-_{3k-1}   \Big \|
    +L_{tU} \epsilon_U \delta_{\tau_U} 
    + \gamma_\Delta
    \label{Eq: DeltaF-4}
\end{equation}
holds for any
$\mathbf{x}_U |^-_{3k-1}  \in B_{l_U}(\mathbf{0})$.

As the robot has full control authority within the OA domain, we can establish a tighter upper bound on $\Big \|   \mathbf{x}_F |^+_{3k} \Big \| $ than Eqs.~\eqref{Eq: DeltaF-3} and \eqref{Eq: DeltaF-4} by applying Proposition 3 from our previous work~\cite{gu2022global}.
That is,
there exists a real, positive number $l_O$ and Lipschitz constants $L_{tO}$ and $L_{xO}$ such that 
\vspace{-0.2 in}
\begin{equation}
    \Big \|   \mathbf{x}_F |^+_{3k} \Big \| 
    \leq 
    L_{tO}
    \sqrt{\tfrac{c_{2O}}{c_{1O}}} 
    e^{-\tfrac{c_{3O}}{2 c_{2O}} \delta_{\tau_O} }
    \Big \| \mathbf{x}_O |^+_{3k-1}   \Big \|
    + 
    L_{xO} 
    \Big \| \mathbf{x}_O |^-_{3k}   \Big \|
    \label{Eq: DeltaF-5}
\end{equation}
for any
$\mathbf{x}_O |^-_{3k}  \in B_{l_O}(\mathbf{0})$.

From Eqs.~\eqref{Eq: ContF-1}, \eqref{Eq: ContF-2}, \eqref{Eq: ContF-3}, and \eqref{Eq: DeltaF-3}-\eqref{Eq: DeltaF-5}, we obtain:
\vspace{-0.2 in}
\begin{equation}
\vspace{-0.2 in}
  \Big \|   \mathbf{x}_F |^+_{3k} \Big \| 
    \leq 
    \bar{N}+\bar{L}\Big \|  \mathbf{x}_F |^+_{3k-3}  \Big \|.
    \label{Eq: FContDelta}
\end{equation}
where
\vspace{-0.2 in}
\begin{equation*}
\vspace{-0.2 in}
\begin{aligned}
\bar{N}:=
&
 \big( 
 L_{tU} \epsilon_U \delta_{\tau_U} + \gamma_\Delta
    + L_{xU} ( \alpha_1  L_{tF} \epsilon_F \delta_{\tau_F} 
    + \alpha_1 \gamma_\Delta
    +\alpha_1 \alpha_2 )
    \big)
    \\
    &\cdot
   \big( 
    L_{tO} 
    \sqrt{\tfrac{c_{2O}}{c_{1O}}} 
    e^{-\tfrac{c_{3O}}{2 c_{2O}} \delta_{\tau_O} }
    + L_{xO} 
    \sqrt{\tfrac{c_{2O}}{c_{1O}}} 
    e^{-\tfrac{c_{3O}}{2 c_{2O}} (1+\epsilon_O)\delta_{\tau_O} }
\big)
\end{aligned}
\end{equation*}
    and
    \vspace{-0.2 in}
    \begin{equation*}
    \vspace{-0.2 in}
        \begin{aligned}
    \bar{L} 
    :=&
L_{xU} 
    \alpha_1 
    L_{xF}
    \sqrt{\tfrac{c_{2F}}{c_{1F}}} 
    e^{-\tfrac{c_{3F}}{2 c_{2F}} (1+\epsilon_F)\delta_{\tau_F} }
    \\
    &\cdot
\big( 
    L_{tO} 
    \sqrt{\tfrac{c_{2O}}{c_{1O}}} 
    e^{-\tfrac{c_{3O}}{2 c_{2O}} \delta_{\tau_O} }
    + L_{xO} 
    \sqrt{\tfrac{c_{2O}}{c_{1O}}} 
    e^{-\tfrac{c_{3O}}{2 c_{2O}} (1+\epsilon_O)\delta_{\tau_O} }
\big),
        \end{aligned}
    \end{equation*}

Using Eqs.~\eqref{Lyap-F-O} and \eqref{Eq: FContDelta}, we obtain
\vspace{-0.2 in}
\begin{equation}
\vspace{-0.2 in}
    V_F|_{3k}^+  
    \leq 
    2 c_{2F} \bar{N}^2
    +
     \tfrac{2c_{2F}\bar{L}^2}{c_{1F}} V_F|_{3k-3}^+ .
     \label{eq: VF-final}
\end{equation}


Note that the scalar positive parameters $\bar{N}$ and $\bar{L}$ in Eq.~\eqref{eq: VF-final} are both dependent on the continuous-phase convergence rates of the Lyapunov-like functions within the OA and FA domains (i.e., $c_{3F}$ and $c_{3O}$),
Specifically, $\bar{N}$ and $\bar{L}$ (and accordingly $\tfrac{2c_{2F}\bar{L}^2}{c_{1F}}$ and $2c_{2F}\bar{N}^2$) will decrease towards zero as the continuous-phase convergence rates increase towards the infinity.

If condition (A3) holds (i.e., the PD gains can be adjusted to ensure a sufficiently high continuous-phase convergence rate), 
we can choose the PD gains such that $\tfrac{2c_{2F}\bar{L}^2}{c_{1F}}$ is less than $1$ and $2c_{2F}\bar{N}^2$ is sufficiently close to $0$, which will then ensure 
$V_F|_{3k}^+  \leq  V_F|_{3k-3}^+$ for any $k\in \{ 1,2,...\}$.

$\hfill
\blacksquare$

\subsection{Proof of Theorem~\ref{thm: 1}}

\label{Appendix: thm1}

By the general stability theory based on multiple Lyapunov functions~\cite{branicky1998multiple}, the origin of the overall hybrid error system described in Eqs.~\eqref{Eq: closed-loop error dynamics} and \eqref{eq: xu dynamics} is locally stable in the sense of Lyapunov if the Lyapunov-like functions $V_F$, $V_O$, and $V_U$ satisfy the following conditions:
\begin{itemize}
    \item [(C1)] The Lyapunov-like functions $V_{F}$ and $V_{O}$ exponentially decrease within the continuous FA and OA phases, respectively.
    \item [(C2)] Within the continuous UA phase, the ``switching-out" value of the Lyapunov-like function $V_{U}$ is bounded above by a positive-definite function of the ``switching-in" value of $V_{U}$; and
    \item [(C3)] The values of each Lyapunov-like functions at their associated ``switching-in" instants form a nonincreasing sequence.
\end{itemize}

If the proposed IO-PD control law satisfies condition (B1), then the control law ensures conditions (C1) and (C2), as established in Proposition 1 and 2, respectively.
By further meeting conditions (B2) and (B3), we know from Proposition 3 that condition (C3) will hold. 
Thus, under conditions (B1)-(B3), the closed-loop control system meets conditions (C1)-(C3), and thus the origin of the overall hybrid error system described in Eqs.~\eqref{Eq: closed-loop error dynamics} and \eqref{Eq: closed-loop error dynamics2} is locally stable in the sense of Lyapunov.
$\hfill
\blacksquare$

\end{document}